\PassOptionsToPackage{table}{xcolor}
\documentclass{article} 
\usepackage[final]{colm2025_conference}
\usepackage{microtype}
\usepackage{url}
\usepackage{booktabs}
\usepackage{lineno}
\usepackage{hyperref}
\definecolor{darkblue}{rgb}{0, 0, 0.5}
\hypersetup{colorlinks=true, citecolor=darkblue, linkcolor=darkblue, urlcolor=darkblue}
\usepackage{tabularx}
\usepackage{enumitem}
\usepackage{graphicx}
\usepackage{times}
\usepackage{latexsym}
\usepackage{floatrow}
\usepackage{caption}
\usepackage{subcaption}
\usepackage{listings}
\usepackage{amsmath}

\definecolor{subtask1}{HTML}{ffc1cd}
\definecolor{subtask2}{HTML}{e3d8ff}
\definecolor{subtask3}{HTML}{aae3ff}
\definecolor{subtask4}{HTML}{c8f8e1}
\definecolor{tax}{HTML}{ffb000}

\usepackage{multirow}
\usepackage{array}
\usepackage{makecell}
\newcolumntype{M}[1]{>{\centering\arraybackslash}m{#1}}

\usepackage{verbatimbox} 
\usepackage{fancyvrb}
\usepackage{listings}
\usepackage{scalefnt}
\usepackage[T1]{fontenc}

\title{LawFlow: Collecting and Simulating Lawyers’ Thought Processes on Business Formation Case Studies}

\author{
\parbox{\linewidth}{
Debarati Das\textsuperscript{1}, Khanh Chi Le\textsuperscript{1}\thanks{Equal Contribution}, Ritik Sachin Parkar\textsuperscript{1}$^*$, Karin De Langis\textsuperscript{1}, \\
Brendan Madson\textsuperscript{2}, Chad M. Berryman\textsuperscript{2}, Robin M. Willis\textsuperscript{2}, Daniel H. Moses\textsuperscript{2}, \\
Brett McDonnell\textsuperscript{2}, Daniel Schwarcz\textsuperscript{2}, Dongyeop Kang\textsuperscript{1}
}\\
\textsuperscript{1}Computer Science and Engineering, University of Minnesota \\
\textsuperscript{2}Law School, University of Minnesota \\
\texttt{\{das00015,bhm,schwarcz,dongyeop\}@umn.edu}\\
}

\begin{document}

\ifcolmsubmission
\linenumbers
\fi
\label{sec:analysis}
\maketitle

\begin{abstract}

Legal practitioners, particularly those early in their careers, face complex, high-stakes tasks that require adaptive, context-sensitive reasoning. While AI holds promise in supporting legal work, current datasets and models are narrowly focused on isolated subtasks and fail to capture the end-to-end decision-making required in real-world practice. To address this gap, we introduce \texttt{LawFlow}, a dataset of complete end-to-end legal workflows collected from trained law students, grounded in real-world business entity formation scenarios. Unlike prior datasets focused on input-output pairs or linear chains of thought, \texttt{LawFlow} captures dynamic, modular, and iterative reasoning processes that reflect the ambiguity, revision, and client-adaptive strategies of legal practice.
Using \texttt{LawFlow}, we compare human and LLM-generated workflows, revealing systematic differences in structure, reasoning flexibility, and plan execution. Human workflows tend to be modular and adaptive, while LLM workflows are more sequential, exhaustive, and less sensitive to downstream implications. Our findings also suggest that legal professionals prefer AI to carry out supportive roles, such as brainstorming, identifying blind spots, and surfacing alternatives, rather than executing complex workflows end-to-end. Our results highlight both the current limitations of LLMs in supporting complex legal workflows and opportunities for developing more collaborative, reasoning-aware legal AI systems. All data and code are available on our project page.\footnote{\url{https://minnesotanlp.github.io/LawFlow-website/}}

\end{abstract}
\section{Introduction}

Legal professionals, especially those early in their careers, face growing pressure to handle increasingly complex tasks - from navigating regulatory compliance to conducting detailed contract negotiations - while managing limited time and resources. Despite the promise of AI in legal domains \citep{schwarcz2025ai,nielsen2024building}, current legal AI solutions remain narrowly scoped. They are often designed for isolated tasks such as contract review or legal research \citep{li2024lecardv2,narendra2024enhancing}, failing to reflect the broader arc of legal work. To realize AI’s potential in legal practice, we argue for a shift in how legal reasoning is represented: not as isolated inputs and outputs, but as full, end-to-end decision processes. 


\begin{figure}[!ht]
  \centering
  \includegraphics[width=0.99\columnwidth,clip, trim=0cm 2.5cm 0cm 0cm]{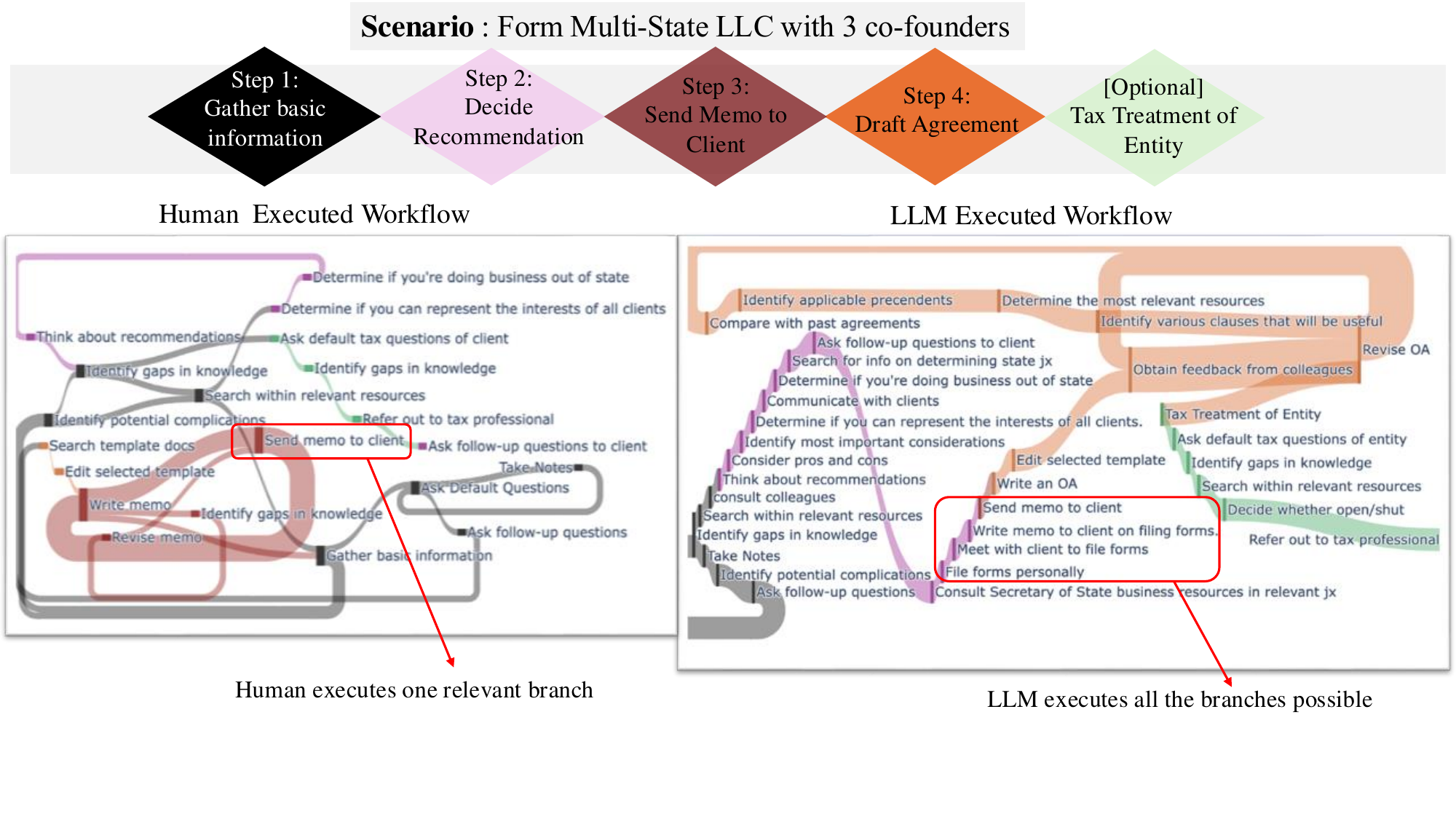}
  \caption{\small \textbf{Comparison of Human (law student) and LLM execution workflows for the same drafting task involving entity formation.} Here, \textit{execution} refers to the selection of the next step in a reasoning process, not task completion, but the model’s decision about what to do next based on context and prior actions. 
  While the human execution is adaptive and iterative, the LLM follows a rigid, linear path, often executing all branches rather than selecting one relevant branch from the task plan. Points of divergence highlight differences in strategy, task decomposition, and assumption handling, revealing key limitations in LLM reasoning and opportunities for complementary human-AI collaboration.}
  \label{fig:main_motivation}
\end{figure}

Capturing legal workflows poses distinct challenges not seen in traditional automation domains. Existing automation and process modeling datasets emphasize deterministic sequences optimized for efficiency. Legal workflows, by contrast, demand interpretive, context-aware reasoning. Practitioners must frequently navigate uncertainty, integrate new information, revise decisions, and balance competing constraints.  To build AI systems that support such work, we propose moving toward modeling not only final outcomes, but also the \textit{processes}, or the ``chain-of-decisions'' that shape them.

Transactional law \citep{goforth2017transactional, feldman1999drafting}, and in particular business entity formation for small commercial ventures, offers a structured yet complex setting in which to study legal reasoning. Drafting an agreement involves eliciting incomplete client information, interpreting legal structures, and adapting language to evolving client goals and regulatory constraints \citep{goforth2017transactional, feldman1999drafting}. While prior datasets focus on static clauses or discrete legal tasks, they do not capture how legal practitioners navigate such workflows from start to finish. 
Legal reasoning in practice is dynamic and non-linear. It involves iterative refinement, contextual judgment, and responsiveness to client needs - patterns that challenge the rigid, task-specific strategies of current AI systems. To address this, we argue that AI must engage with the reasoning processes that underlie legal decision-making, not just the surface-level outputs. Traditional chain-of-thought (CoT) reasoning datasets \citep{wei2022chain, kim2023cot} are insufficient for this goal. They typically model reasoning as linear, single-turn problems with a single correct answer. Legal workflows, by contrast, require what we term ``chain-of-decisions reasoning'': multi-turn, context-aware processes that involve ambiguity, revision, and multiple plausible paths forward \citep{wang2024drivecot}. This richer mode of reasoning better reflects the realities of legal practice, where flexibility and adaptation are essential.

To support this shift, we introduce \texttt{LawFlow}, a dataset that captures full end-to-end legal workflows from trained law students working through realistic business formation scenarios. By tracing the full decision arc - from client intake to operating agreement drafting - \texttt{LawFlow} enables fine-grained analysis of how human and LLM-generated 
workflows differ in structure, execution, and adaptability.

The chain-of-decisions reasoning captured in \texttt{LawFlow} allows for comparisons between human and LLM-generated workflows. For example, we can identify \textit{points of divergence}, where strategies, task decomposition, or assumptions differ (highlighted in Figure \ref{fig:main_motivation}). These divergences reveal limitations in current LLMs such as difficulties handling ambiguity, anticipating downstream implications, or revising incomplete plans which are failures that are not visible through traditional CoT evaluation alone. However, emerging reasoning-focused models \citep{guo2025deepseek, jaech2024openai} show promise in addressing some of these issues, particularly in maintaining coherence across steps and accounting for long-range dependencies. Studying these differences also raises the possibility of mutual improvement: how might AI learn from human flexibility, and how might humans benefit from AI’s structure or recall?


\texttt{LawFlow} supports research on key questions at the intersection of legal reasoning and LLMs: \begin{enumerate}[noitemsep, topsep=0pt, left=1em] \item How do human and LLM-generated legal workflows differ in structure, execution, and adherence to task plans? \item Can we model diverse legal workflows and flag unlikely or anomalous steps, without presuming a single correct path? \item What key decision points shape legal workflows, and how might AI systems support users at these junctures? \end{enumerate}
\textbf{Main Findings.} The fine-grained procedural data captured by \texttt{LawFlow} provides insight into these questions. Specifically, we find that human workflows tend to be more modular and adaptive, with subtasks often revisited or reordered based on new information. In contrast, LLM-generated workflows typically follow a rigid, linear sequence, with limited revision or backtracking. These findings reinforce the need for AI systems that support flexible, context-sensitive reasoning. 
\begin{figure}[t!]
  \centering
  \includegraphics[width=0.9\columnwidth,clip, trim=2cm 2.5cm 6.5cm 2.5cm]{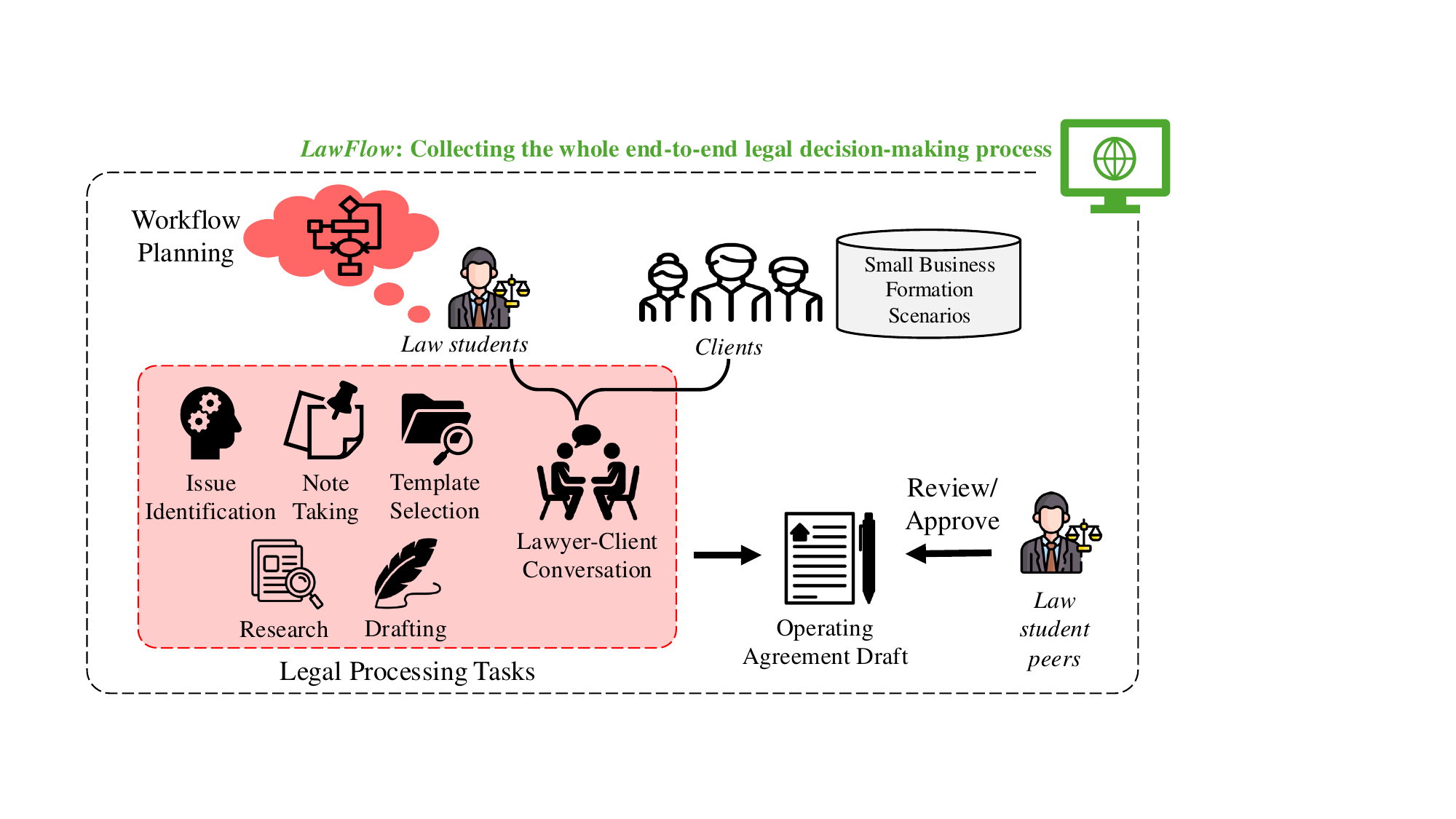}
  \caption{ \small \textbf{Overview of the \texttt{LawFlow} dataset creation}, which captures novice law practioner decision-making in drafting operating agreements for small business formation. Starting with realistic business formation scenarios, law students and clients simulate authentic client-information elicitation sessions. This is followed by sub-processes such as issue identification, note-taking, legal research, template selection, and drafting, with every human decision and corresponding reasoning recorded by the \texttt{LawFlow} data collection tool. The final operating agreements and memos are then reviewed and verified by other law students, emulating the oversight of senior partners in a law firm.}
  \label{fig:motivational_figure}
\end{figure}
\section{Background}
\label{sec: Related work}
\textbf{Lack of focus on legal workflows}. Much of the current progress in legal AI has centered on developing models tailored to narrow, well-scoped tasks such as legal classification \citep{lee2023lexgpt}, judgment prediction \citep{sesodia2025annocaselaw, gray2024using}, contract comparison \citep{narendra2024enhancing}, and legal research \citep{li2024lecardv2, li2023sailer}. These tasks are attractive because they are easier to formalize, enabling the construction of annotated datasets that support targeted evaluation and fine-tuning. However, while such datasets have accelerated progress in individual subtasks, they fall short of capturing real-world legal practice's complex, multistep nature. Studies show that LLMs perform well on classification-style tasks but struggle with application-oriented challenges that require procedural reasoning and contextual understanding \citep{guha2023legalbench, li2024lexeval}. This gap is particularly problematic because legal workflows consist of steps such as building arguments, coordinating filings, or interpreting case metadata, which are not just procedural but inherently human-centric. Lawyering involves iterative decision-making, ethical judgment, and value-laden trade-offs that static benchmarks cannot fully capture \citep{choi2024lawyering}. As AI reaches or exceeds human performance on existing tests, these benchmarks are becoming saturated \citep{artificiallawyer2024}
and increasingly inadequate for evaluating legal AI's true utility. While task-oriented AI already provides valuable support in legal settings \citep{schwarcz2025ai}, there is growing interest in exploring how such capabilities might be integrated into the broader workflows of legal professionals. \textit{Bridging this gap requires a new focus on modeling legal workflows, dynamic, structured, and human-in-the-loop processes that better reflect the realities of legal decision-making.}

\begin{figure}[t!]
  \centering
  \includegraphics[width=1\columnwidth,clip, trim=0cm 0cm 0cm 0cm]{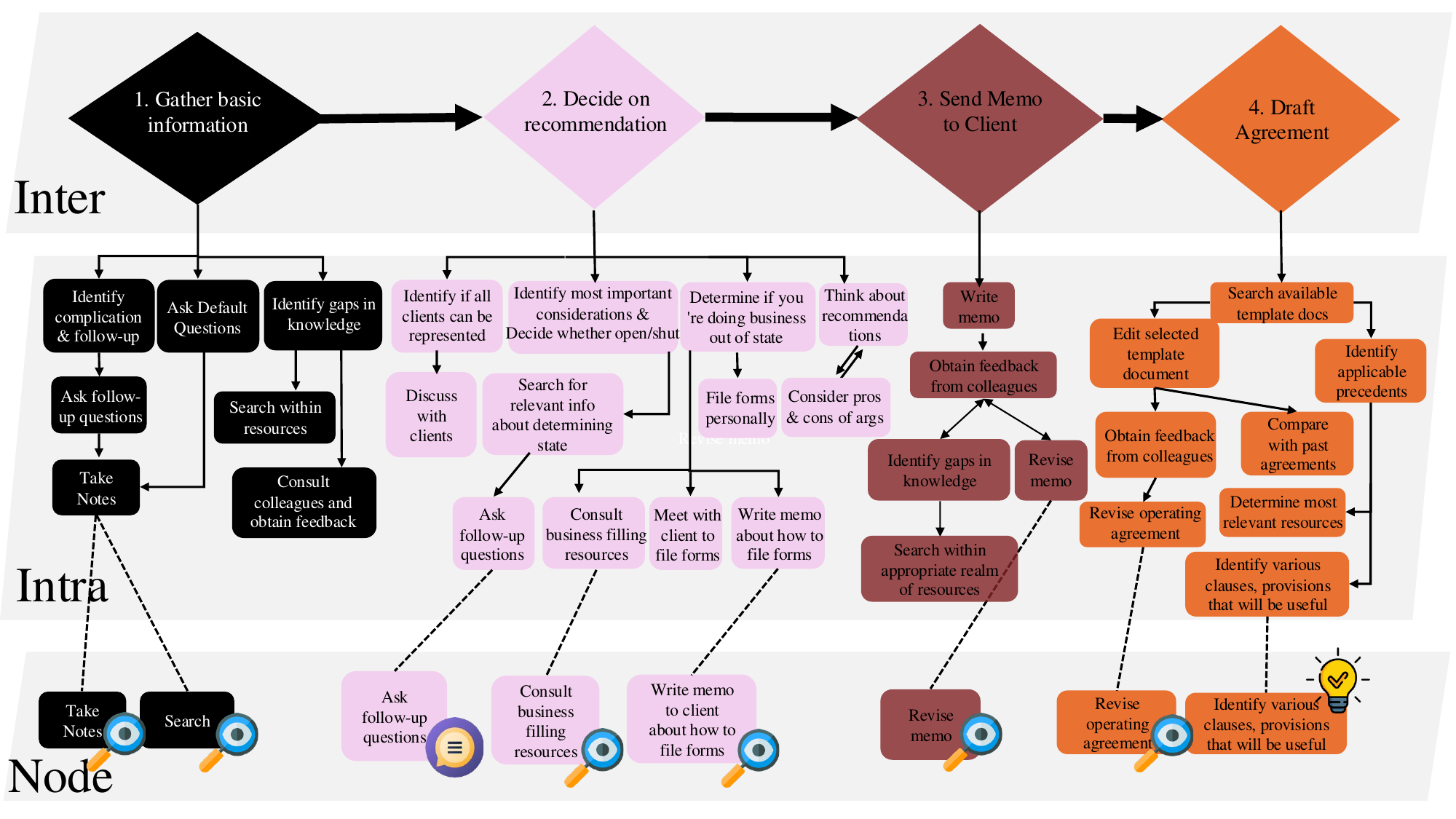}
  \caption{\small \textbf{Expert-informed task diagram for business entity formation.} This diagram illustrates the multi-level structure of the legal workflow behind drafting an Operating Agreement. The inter-subtask level (top row) outlines the major stages of the process, from initial client intake to final document preparation. The intra-subtask level (middle row) decomposes each stage into finer-grained tasks and shows their interconnections. The node-level (bottom row) represents individual actions, annotated by cognitive modality: introspective (internal legal reasoning), interactive (client or colleague communication), and observable (use of external tools and resources). Thus, the task diagram captures the complexity, adaptivity, and tool-mediated nature of real-world legal reasoning.}
  \label{fig:task_diagram}
\end{figure}

\textbf{Inability of AI to successfully capture legal workflows} Although workflows have been successfully modeled in domains like enterprise automation through Robotic Process Automation (RPA) \citep{wewerka2020robotic, ferreira2020evaluation}, these systems rely on hand-crafted rules and are limited to repetitive, well-structured tasks. Legal workflows, by contrast, are inherently domain-specific and human-centric—they require ongoing interpretation and adaptation to evolving facts and norms. Unlike deterministic enterprise settings, legal practice involves ambiguity, negotiation, and contextual judgment, which cannot be captured through rigid scripts or predefined templates \citep{herm2020consolidated}. Recent efforts in Autonomous Process Automation (APA) have explored using large language models to structure workflows dynamically \citep{ye2023proagent, fan2024workflowllm}, showing promise in domains like travel planning and enterprise operations \citep{xie2024travelplanner, wornow2024automating}. However, even state-of-the-art models like GPT-4 struggle with orchestrating complex, multi-step processes that require domain knowledge and procedural nuance \citep{zeng2023flowmind}. These models tend to oversimplify workflows, failing to maintain coherence across steps or account for the interpretive reasoning central to legal decision-making. This highlights a core limitation: \textit{despite advances in general workflow automation, AI remains unable to manage domain-specific workflows that demand flexible, expert-level reasoning.}

\textbf{From Chain-of-Thought to Chain-of-Decisions: A Shift in Reasoning Paradigms} Existing chain-of-thought (CoT) datasets \citep{wei2022chain, kim2023cot} are primarily designed around single-turn, single-task reasoning, where a problem is decomposed into a static sequence of logical steps culminating in a single, objectively correct answer. These datasets have been valuable for training models to break down complex problems into interpretable sub-steps \citep{wang2024openr, guo2025deepseek}, but they offer limited utility when it comes to modeling real-world decision-making processes. In contrast, domains like legal practice demand reasoning over “chain-of-decisions” workflows - dynamic, multi-turn processes that evolve over time, depend on shifting contexts, and often involve multiple valid paths forward. These workflows are not only shaped by legal rules but also by client-specific goals, risk tolerance, and strategic judgments, often requiring backtracking, revision, and branching. As such, methods trained exclusively on CoT-style data, which assume linearity and objective correctness, fall short in capturing the complexity, subjectivity, and flexibility inherent in expert-driven workflows \citep{chen2024huatuogpt, wang2024drivecot}. This reveals a key challenge: \textit{current CoT datasets and reasoning paradigms are ill-suited for domains that require nuanced, iterative chains of decisions rather than fixed chains of thought.}
Recent studies that record scholars' entire writing process through keystroke logs \cite{wang2025scholawrite,koo2023decoding} have motivated our approach. Unlike scientific writing, however, legal processing data involves more than drafting documents, such as note-taking or preparing operating agreements, it also requires deliberate, high-level reasoning and decision-making steps that must be explicitly planned.

\section {LawFlow Dataset Creation}

\label{sec:expert-task-diagram}
\textbf{Human Task Plan Construction}.
We conducted in-depth interviews with senior law faculty and law students to understand how lawyers handle business entity formation towards drafting an Operating Agreement and modeled the entire end-to-end workflow. The resulting expert-informed task plan(shown in Figure~\ref{fig:task_diagram}) captures this process across three levels of granularity: \textbf{inter-subtask} (the major workflow stages), \textbf{intra-subtask} (the interconnected steps within each stage), and \textbf{node-level} (individual actions annotated by cognitive modality). These modalities include introspective (individual reasoning), interactive (communication with clients or colleagues), and observable (engagement with tools or resources). The workflow reflects real-world legal practice, spanning information gathering, strategic decision-making, client communication, and iterative document drafting while highlighting how law students coordinate tools, knowledge, and judgment throughout. The main subtasks involved in this task plan are : 
\begin{enumerate}[noitemsep, topsep=0pt, left=1em]
    \item \textit{Information Gathering:} The lawyer elicits basic information from the client, identifies potential legal or factual complexities, takes notes, and formulates follow-up questions. This subtask involves active client interaction and may require dynamic adaption by the lawyer, who might need to consult external resources to address jurisdiction-specific issues or unusual structures.
    \item \textit{Deciding a Recommendation:} Based on gathered information, the lawyer assesses the legal and strategic dimensions of the case. This includes determining jurisdictional applicability, conducting further research if necessary, and deciding whether to proceed with a legal recommendation.The lawyer must also assess whether they can adequately represent the client’s interests and might schedule an additional follow-up. 
    \item \textit{Drafting and Sending a Memo:} The lawyer formalizes their analysis and recommendation in a client-facing memo. The memo outlines the proposed structure and identifies areas of uncertainty or risk. It may be revised based on feedback from colleagues or newly surfaced information.
    \item \textit{Drafting the Operating Agreement (OA):} Using prior templates or precedent documents, the lawyer drafts the OA, tailoring clauses to the client’s needs. This iterative process involves comparative analysis, peer feedback, and revision for legal clarity and client alignment.
    \item \textit{Assessing Tax Treatment [Optional]:} In scenarios where tax implications are relevant, the lawyer gathers tax-related information, conducts targeted legal research, and may refer the client to a tax specialist. This subtask is selectively activated based on the complexity of the client’s needs. Note that this subtask is not highlighted in Figure~\ref{fig:task_diagram} because of space constraints but is described in the Appendix figure \ref{fig:memo}.
\end{enumerate}

\textbf{Simulation of Legal Workflow Scenarios}.
\label{sec:seed-scenario}
To contextualize our task decomposition and analyze entity formation workflows, we focus on 3rd year high-performing law students with experience in business law - representative of the novice practitioners that educational and assistive AI systems aim to support through scaffolded reasoning and task guidance rather than expert replacement. One of the participating law students created a set of realistic seed scenarios. 

These scenarios were based on anonymized real-world cases from an affiliated legal clinic. Each scenario poses a unique small business formation challenge, crafted to surface varying legal considerations, client needs, and decision-making points. To simulate realistic legal interactions, we conducted structured roleplays where law students play-acted as lawyers advising clients. Computer science students were assigned the role of clients, enabling rich, interactive simulations of the legal process. During these roleplays, law students used the LawFlow Tool (described in Section~\ref{sec:lawflow-tool}) to document their reasoning, decisions, and actions. Each action was tagged with the corresponding subtask from our expert-informed legal task plan, allowing us to trace how legal workflows unfold in real-time.

Each scenario is annotated with two metadata types-complexity and nuance parameters (Appendix \ref{sec:client_elicitation}), which capture structural and interpersonal factors that influence legal task execution. While complexity parameters affect structural aspects of the legal task, such as workflow scope and subtask emphasis, nuance parameters shape the realism and interactive dynamics of the roleplay, enabling analysis of how different contexts impact legal reasoning.
\begin{figure}[t!]
  \centering
  \vspace{-3.5mm}
  \includegraphics[width=0.950\columnwidth,clip, trim=0cm 0cm 0cm 0cm]{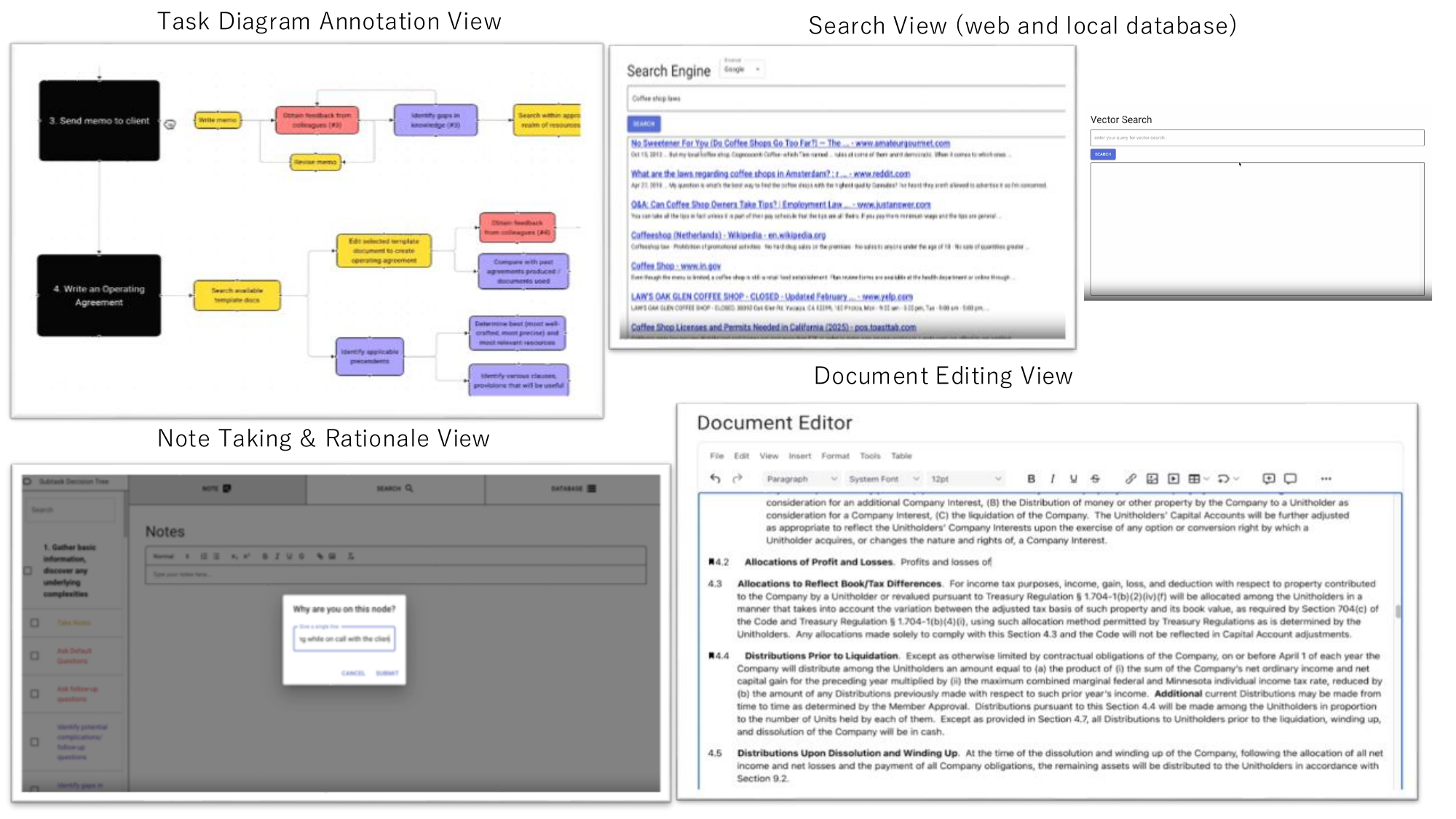}
  \caption{\small \textbf{Features of the LawFlow Collection Tool} include note taking, document editing, search as well as task diagram annotation.}
  \label{fig:tool_capture}
\end{figure}

\textbf{LawFlow Data Collection Tool}.
\label{sec:lawflow-tool}
To support data collection during legal roleplays, we developed a web-based application designed to reflect key elements of a law student’s workflow during entity formation. Figure \ref{fig:tool_capture} showcases the different views of this web application. The tool includes several core features:
\begin{itemize} [noitemsep, topsep=0pt, left=1em]
\item \textit{Subtask Annotation}: law students or the users of this tool, can tag each action with a relevant node from the expert task plan. Subtasks are non-sequential and optional, allowing for naturalistic variation in how people approach the task. Users can also annotate their decision-making rationale alongside each tag.
\item \textit{Note Taking}: A lightweight, persistent text field allows users to take informal notes during interactions (e.g., client intake), with full keystroke logging for fine-grained analysis.
\item \textit{Template Library}: Users can access a curated set of Agreement templates, simulating real-world use of precedents. Search activity is logged to capture document reuse behavior.
\item \textit{Legal Search Interface}: Users can conduct open web searches and template-specific searches, simulating the use of proprietary legal research tools and allowing us to study how search behavior influences decision-making.
\item \textit{Document Editing}: A built-in word processor (powered by TinyMCE \footnote{\url{https://github.com/tinymce/tinymce}}) supports legal drafting. All keystrokes are logged, enabling reconstruction and analysis of the drafting process.
\end{itemize}
These features enable comprehensive capture of legal decision-making, tool usage, and workflow adaptation. Thus, the LawFlow Data Collection Tool allows us to observe what decisions are made and how law students navigate uncertainty, balance competing constraints, and structure their reasoning in practice.

\textbf{Dataset of Legal workflows}.
The entity formation workflow typically produces two key artifacts: an \textit{Agreement} and, optionally, a \textit{client memo}. The Agreement is generated by selecting a template from a curated library and customizing it to fit the client’s business structure, goals, and legal needs, through clause additions, removals, or edits informed by legal reasoning and client input. The client memo serves as a communication tool, explaining key decisions, legal obligations, and any open issues. Together, these documents reflect both the technical drafting and the student’s interpretive reasoning. 

Our collected dataset includes 10 finalized scenarios. While 20 were initially collected, the first 10 were excluded due to task plan revisions and RA acclimation to the tool. The final dataset covers diverse legal contexts (10 memos and 8 agreements) across multiple entity types like LLCs, non-profits etc). In 2 cases, the RA opted not to draft an agreement due to client representation conflicts or limited expertise, which are realistic outcomes reflecting legal ethics and complexity. Sessions lasted an average of 101 minutes. Appendix Tables \ref{tab:scenario-diversity} and \ref{tab:business-entity-metric} illustrate the complexity and diversity in our dataset. While our main focus was on human and LLM workflows, we also evaluated output agreement quality through human and LLM reviews. To support this, experts compared LLM and human drafts for clarity, coverage, and issue-spotting. In some cases, LLMs were preferred or surfaced overlooked issues, showing potential as second-pass reviewers. These findings are detailed in the Section \ref{sec:eval_quality}.

\section {Analysis and Discussion} 
  \label{sec:sec4}
This section presents our experimental framework and presents results for comparing how law students (\textcolor{orange}{humans}) and \textcolor{green!50!black}{LLMs} plan and execute the full end-to-end legal workflows, instead of a single task execution. We focus on three main areas: (\ref{sec:rq1}) identifying structural and behavioral differences between human and AI workflows, (\ref{sec:rq2}) examining the diversity and monitoring of legitimate workflow paths, and (\ref{sec:rq3}) pinpointing critical decision moments where AI might assist.

\textbf{Construction of Plans and Execution Graphs}. 
To conduct our analysis, we collected structured task plans and execution traces from both law students and LLMs. These plans and execution traces can be visualized as graphs with nodes representing the subtasks and edges representing the transitions between them. Senior law faculty and law students helped develop an ``near-ideal'' \textit{Human Task Plan} (Section \ref{sec:expert-task-diagram}), while law students participated in scenario-based roleplays using the LawFlow data collection tool, which logged their actions and decisions. 

From these logs, we constructed a \textit{Human Execution Graph}, Deviations from the Human Plan highlight reordering, omission, or improvisation. As SOTA reasoning models are known to be trained to be effective for multi-step planning, we prompted reasoning-focused LLMs like GPT-O1 and Deepseek-R1 \citep{jaech2024openai,guo2025deepseek} with the same scenarios to generate two outputs: 
(1) \textit{LLM Task Plan} – A high-level plan generated from multiple business scenarios. It aims to capture a broader strategy that covers various use cases, rather than a single scenario. 
(2) \textit{LLM Execution Graph} – A trace of decisions about which task to perform next in a given scenario. Here, \textit{execution} refers not to carrying out a task, but to the model’s decision about what the next step should be, given the current context, prior actions, and scenario-specific details. Each step simulates a legal reasoning move-either a hypothetical client interaction or internal deliberation, followed by a next-step decision. The executed steps are then collected to form the \textit{LLM Execution Graph}. Prompts and outputs for these steps are detailed in \ref{sec:llm_workflow_prompt}. This representation differs from traditional chain-of-thought (CoT) reasoning, which typically involves linear, forward-only justifications toward a known answer. Instead, our graphs capture chain-of-decisions reasoning: multi-turn, context-aware navigation through a branching task space, often involving revision, ambiguity, and multiple plausible paths.
\subsection {How do human and LLM-generated legal workflows differ in structure, execution, and adherence to plans?}\label{sec:rq1}
We investigate how legal workflows differ when generated and executed by law students versus LLMs, focusing on three dimensions: (1) the structural characteristics of task plans, (2) adherence to those plans during execution, and (3) behavioral patterns during task performance.

\begin{figure}[tbp]
  \centering
  \begin{minipage}[c]{0.48\textwidth}
    \small
    \setlength{\tabcolsep}{3pt}
    \begin{tabular}{>{\raggedright\arraybackslash}p{3cm}cc}
        \textbf{Graph Metrics} & \textbf{Human Plan} & \textbf{LLM Plan} \\ 
        \midrule
        Tree Depth & 3.6 ± 0.5 $\uparrow$ & 1.0 ± 0.0 \\  
        \midrule
        Number of Subtasks & 5 $\downarrow$ & 10 \\  
        \midrule
        Branches/Subtask & 4.8 ± 2.3 $\uparrow$ & 3.8 ± 0.4\\  
        \midrule
        Nodes/Subtask & 8.8 ± 2.6 $\uparrow$ & 3.8 ± 0.4\\  
        \bottomrule
    \end{tabular}
    \vspace{2mm}
    {\par\small\textbf{(a)} Humans plan hierarchically, starting with a few high-level subtasks (number of subtasks) and expanding them into deeper, more detailed structures (tree depth and branches/subtask). In contrast, LLMs generate flatter, sequential plans with more top-level subtasks but less internal organization and modularity.}
  \label{tab:human_vs_llm_metrics}
  \end{minipage}
  \hfill
  \begin{minipage}[c]{0.48\textwidth}
    \includegraphics[width=\linewidth]{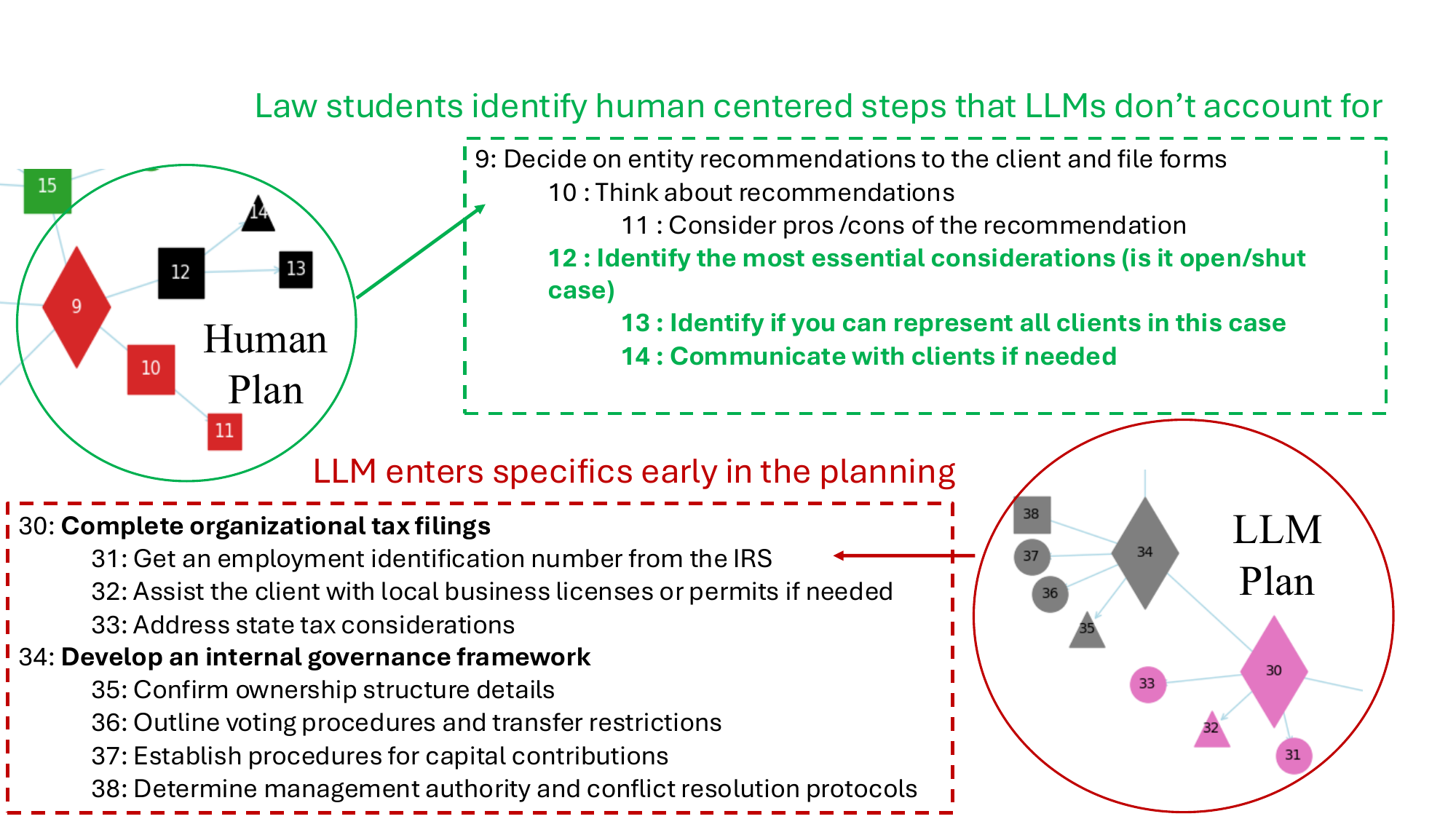}
    \vspace{-2mm}
    {\par\small\textbf{(b)} The figure shows LLM task plan nodes colored by subtask, and the human task plan is mapped to the LLM task plan based on semantic similarity of subtasks. We can see multiple LLM subtasks correspond to the same human subtask. Black nodes indicate nodes which remain unmapped in the human plan (indicating that there are some human tasks LLMs do not account for).}
  \label{fig:human_vs_llm_diagram}
  \end{minipage}
  \caption{\small \textbf{Difference between \textcolor{orange}{Human} and \textcolor{green!50!black}{LLM} Plans.} Humans plan hierarchically with few high-level subtasks and deeper trees. LLMs use flatter structures with more top-level subtasks but lower internal organization.}
  \label{fig:human_vs_llm_combined}
\end{figure}

\paragraph{Task Plan Structure.}
We want to understand the differences between humans and LLMs
in planning out legal workflows and what those differences look like structurally. In service of that, we explore the modularity and branching of their task plans via the following graph-based metrics: \textit{Average tree depth} (the number of hierarchical layers in the task plan), \textit{Number of top-level subtasks} (reflecting the initial modular structuring), \textit{Average branches per subtask} (capturing the extent of decomposition), and \textit{Average nodes per subtask} (measuring the level of internal detail).

Our results indicate that \textcolor{orange}{human task plans are modular and hierarchical}, often beginning with a few high-level subtasks that are \textcolor{orange}{incrementally decomposed into finer-grained actions} (Figure \ref{fig:human_vs_llm_combined}(b)). In contrast, LLM-generated plans are \textcolor{green!50!black}{flatter and more linear, typically enumerating long lists of actions without deeper structural organization}. Table (a) in Figure 5 shows that human plans exhibit greater tree depth (3.6 vs. 1.0) and more nodes per subtask (8.8 vs. 3.8), indicating deeper and more layered decompositions. LLM plans, on the other hand, contain more top-level subtasks (10 vs. 5), reflecting a breadth-first approach with less internal depth. As illustrated in the figure, multiple LLM subtasks often map to a single subtask in the human plan, indicating coarser granularity. Black nodes in the human plan represent tasks not captured in the LLM plan, for example ``consulting colleagues and obtaining feedback'' or ``identifying gaps in knowledge'' are common human steps which are not accounted for by the LLM.  Details about the nodes in the figure can be found in the Appendix Figure \ref{fig:human_llm_plan}. These patterns reinforce a key distinction: \textit{human plans emphasize modularity and depth, while LLM plans prioritize surface-level breadth}.

\paragraph{Task Plan Adherence.}

We also investigate how closely each agent, human or LLM, follows its own plan during execution. This helps us understand not just how workflows are formulated, but how faithfully they are carried out in practice. 
To quantify adherence, we use two complementary metrics: \textit{Node Execution Rate} (the proportion of planned actions that were actually executed) and \textit{Levenshtein Distance} (the textual similarity between the linearized plan graph and execution graph; lower values indicate closer alignment).

Table~\ref{tab:lhs_exec} compares how closely each agent follows its own plan. \textcolor{green!50!black}{LLMs tend to stick more closely to their original task plans}, with a higher node execution rate (0.78 vs. 0.54) and lower Levenshtein distance between planned and executed sequences. This suggests a more \textcolor{green!50!black}{rigid and deterministic execution style}. Human executions show greater deviation, consistent with the intuition of a \textcolor{orange}{more adaptive workflow that involves reordering, skipping, or refining tasks in response to real-time judgment}.

\begin{table}[tbp]
\centering
\small
\begin{subtable}[t]{0.48\textwidth}
    \centering
    \small
    \begin{tabular}{p{0.3\linewidth}@{}p{0.3\linewidth}@{}p{0.35\linewidth}}
      \textbf{Metrics} & \centering\arraybackslash \makecell{\textbf{Human Exec} \\(Human Plan)} & \centering\arraybackslash \makecell{\textbf{LLM Exec} \\(LLM Plan)} \\
      \midrule
      Node exec rate & \centering\arraybackslash 0.53 ± 0.29 & \centering\arraybackslash 0.78 ± 0.14 $\uparrow$\\
      \midrule
      Levenshtein dist. & \centering\arraybackslash 49.30 ± 19.59 & \centering\arraybackslash \begin{tabular}{@{}l@{}}10.70 ± 6.53 (before) \\ 37.90 ± 0.74 (after) $\downarrow$\end{tabular} \\
      \bottomrule
    \end{tabular}
    \caption{\small \textbf{Adherence to Task Plan}. LLMs exhibit higher adherence to their own task plans than humans, with higher node execution rates and lower textual deviation. We also align the LLM-execution with the human execution, thus (before) shows the scores pre-mapping and (after) shows the scores post-mapping.}
    \label{tab:lhs_exec}
\end{subtable}
\hfill
\begin{subtable}[t]{0.48\textwidth}
    \centering
    \small
    \begin{tabular}{p{0.4\linewidth}@{}p{0.3\linewidth}p{0.25\linewidth}}
      \textbf{Metrics} & \centering\arraybackslash \makecell{\textbf{Human Exec} \\(Human Plan)} & \centering\arraybackslash \makecell{\textbf{LLM Exec} \\(Human Plan)} \\
      \midrule
      Number of cycles & \centering\arraybackslash 13.40 ± 16.24 $\uparrow$ & \centering\arraybackslash 0.40 ± 0.70 \\  
      \midrule
      Visit count & \centering\arraybackslash 2.18 ± 0.65 $\uparrow$ & \centering\arraybackslash 1.04 ± 0.05\\  
      \midrule
      Subtask transition  & \centering\arraybackslash 12.80 ± 6.12 $\uparrow$ & \centering\arraybackslash 5.30 ± 0.48 \\  
      \midrule
      Node execution rate & \centering\arraybackslash 0.53 ± 0.29 & \centering\arraybackslash 0.77 ± 0.04 $\uparrow$\\  
      \bottomrule
    \end{tabular}
    \caption{\small \textbf{Execution Patterns}. Humans execute plans non-linearly with cycles and frequent transitions; LLMs execute linearly with fewer deviations.}
    \label{tab:rhs_exec}
\end{subtable}
\caption{\small Comparison between \textcolor{orange}{Human} and \textcolor{green!50!black}{LLM} execution of task plans, across human- and LLM-generated plans.}
\label{tab:combined}
\end{table}

\paragraph{Structure-based Execution Behavior.}
Understanding how agents (human or LLM) carry out legal work is just as important as how they plan it. While structural differences in task plans reveal how humans and LLMs conceptualize workflows, their execution behavior sheds light on the reasoning strategies they employ in practice—whether they follow a fixed linear path, adapt in real-time, or select steps dynamically. To investigate this, we compute the following metrics to capture these execution dynamics: \textit{Number of cycles} (how often tasks are revisited), \textit{Average visit count per node} (frequency of re-engagement with individual steps), \textit{Subtask group transition count} (how often execution shifts between subtask categories), and \textit{Node execution rate} (repeated here to connect execution behavior to plan coverage).

When executing the same expert-defined plan, \textcolor{orange}{humans exhibit highly dynamic behavior}. As shown in Table~\ref{tab:rhs_exec}, they revisit subtasks frequently (17.5 cycles), re-engage individual steps (2.3 average visits), and shift between subtask categories (13.2 transitions), suggesting \textcolor{orange}{non-linear, adaptive execution}. LLMs, by contrast, \textcolor{green!50!black}{show no cycles, visit each node only once (1.0), and exhibit fewer transitions between subtask categories}. Their node execution rate is also higher (0.8 vs. 0.5), reflecting \textcolor{green!50!black}{more exhaustive coverage} of the original plan, but with less variation in execution flow.

\textbf{Granular Assessment}.
To further characterize these behaviors, we analyze execution at three levels in Figure \ref{fig:task_diagram}:
\begin{itemize}[noitemsep, topsep=0pt, left=1em]
\item \textit{Node Level} - We examine how specific tasks are performed, focusing on ``selecting a business entity type'', which has significant downstream implications for the final agreement. We measure the match rate, or how often LLM choices align with those made by humans.
\item \textit{Intra-subtask Level} - This level assesses how thoroughly agents complete the steps within each subtask. We use a coverage metric to check whether all expected actions (as defined in the human plan) are carried out. We also calculate the Levenshtein distance between each agent’s execution and the human plan to measure how much their workflow diverges through edits.
\item \textit{Inter-subtask Level} - We evaluate how agents transition between subtasks. Using the Longest Common Subsequence (LCS), we analyze the presence of cycles or deviations in task order. We also compute the total number of edits (e.g., insertions) across scenarios to determine which agent alters the planned sequence more frequently.
\end{itemize}

In Table \ref{tab:granularity}, we observe that at the node level, the LLMs achieved 60-70\% match rate with humans in entity selection, after the execution of the client elicitation stage. This could have implications on the final agreement drafted. At the intra-subtask level, the coverage metric indicates that \textcolor{orange}{humans prioritized early-stage actions such as client intake} (0.74±0.28), often skipping lower-yield steps. \textcolor{green!50!black}{LLMs, meanwhile, allocated effort more uniformly across subtasks}, with high emphasis on drafting the operating agreement (0.96±0.06) and tax considerations (0.85±0.05). However, since most scenarios in our roleplay simulations involved minimal tax complexity, which is an artifact of our simulation, the \textcolor{orange}{lower human engagement with tax tasks (0.39±0.44) may reflect greater context sensitivity}. Humans also showed \textcolor{orange}{more iterative execution}, with higher Levenshtein distances between plan and execution (52.9±24.8), suggesting more frequent reordering or revision. At the inter-subtask level, \textcolor{orange}{human workflows commonly included return cycles and multiple edits}, whereas LLMs followed \textcolor{green!50!black}{strictly sequential paths}. Overall, we observe that reasoning models like Deepseek-R1 aligns more with human intra-level behavior (lower coverage on subtask 2 and 5). 

\newcommand{\blackcell}[1]{\cellcolor{black}{\color{white}#1}}

\definecolor{lightpink}{RGB}{255, 228, 240}

\definecolor{lightbrown}{RGB}{222, 184, 135}

\definecolor{lightorange}{RGB}{255, 165, 0}

\definecolor{lightgreen}{RGB}{210, 245, 210}

\begin{table}[t]
    \centering
    \setlength{\arrayrulewidth}{0.9pt}
    \renewcommand{\arraystretch}{1.4}
    \scriptsize
    \resizebox{\textwidth}{!}{%
  
    \begin{tabular}{p{0.2cm}|p{0.2cm}|p{2.88cm}|p{1.2cm}|p{1.5cm}@{}p{1.5cm}|p{6.2cm}}
    \multicolumn{1}{c}{} & 
    \multicolumn{1}{c}{} & 
    \multicolumn{1}{c}{\textbf{Metrics}} & 
    \multicolumn{1}{c}{\textbf{Human exec}} & 
    \multicolumn{1}{c}{\textbf{O1 exec}} & 
    \multicolumn{1}{c}{\textbf{R1 exec}} & 
    \multicolumn{1}{c}{\textbf{Inference}} \\
        \midrule
         \parbox[t]{2mm}{\rotatebox[origin=c]{90}{\small{Node}}}
        & \multicolumn{2}{c|}{\makecell{\small{Match Rate} of \\Business Entity Formation}} & 
        -- & 
        70\%  & 60\%  & \small{LLMs and Humans Disagree on Formed Entities} \\
        \midrule
         \parbox[t]{2mm}{\multirow{12}{*}{\rotatebox[origin=c]{90}{\small{Intra}}}}
        & 
        \parbox[t]{2mm}{\multirow{12}{*}{\rotatebox[origin=c]{90}{\small{Coverage}}}}
        & \blackcell{1. Gather client info} & 0.74 ± 0.28 & 0.69 ± 0.21 & 0.44 ± 0.17 & \small{Key step for humans with high coverage of all tasks} \\ 
        \cline{3-7}
        & & \cellcolor{lightpink}{2. Decide Recommendation} & 0.63 ± 0.31 & 0.84 ± 0.17 & 0.31 ± 0.17 &  \small{Not so key step for humans with many steps skipped compared to O1. R1 differs.} \\ 
        \cline{3-7}
        & & \cellcolor{lightbrown}{3. Draft Memo} & 0.45 ± 0.33 & 0.37 ± 0.32 & 0.53 ± 0.41 & \small{Low human and LLM coverage suggests some subtasks are unnecessary.}\\ 
        \cline{3-7}
        & & \cellcolor{lightorange}{4. Draft Agreement} & 0.40 ± 0.32 & 0.96 ± 0.06 & 0.68 ± 0.24 & \small{Humans cover fewer nodes; task plan may be over-specified.} \\ 
        \cline{3-7}
        & & \cellcolor{lightgreen}{5. Tax Treat} [Optional] & 0.39 ± 0.44 & 0.85 ± 0.05 & 0.19 ± 0.17 & \small{O1 over-covers tax; humans skip due to contextual simplicity, as does R1} \\ 
        
        
        \midrule
    \parbox[t]{2mm}{\rotatebox[origin=c]{90}{\small{Inter}}}
        & \multicolumn{2}{c|}{\small{Longest Common Sequence}} & 
        \tiny{
        \begin{tabular}{@{}l@{}} 1$\rightarrow$2 (80\%) \\ 1$\rightarrow$2$\rightarrow$1 (30\%) \end{tabular} }
        & 
        \tiny{1$\rightarrow$2$\rightarrow$3$\rightarrow$4$\rightarrow$5 (100\%)} 
        & \tiny{\begin{tabular}{@{}l@{}} 1$\rightarrow$2 (80\%) \\ 1$\rightarrow$2$\rightarrow$3$\rightarrow$4 (40\%) \\ 1$\rightarrow$2$\rightarrow$3$\rightarrow$4$\rightarrow$5 \\(20\%)\end{tabular}}
        & \vspace{-5mm}\small{LLM strictly follows the plan, while humans deviate based on the scenario, with no consistent pattern across projects or law RAs.}\\
        \midrule
    \end{tabular}%
    }
    \caption{\small \textbf{Analysis across different granularities} shows that human workflows show adaptive, context-sensitive behavior with iterative revisions and nonlinear transitions, while LLMs exhibit rigid, sequential execution with uniform but less context-aware task coverage.} 
    \label{tab:granularity}
\end{table}


  \vspace{-2mm}
\begin{figure}[htbp]
  \centering

  \begin{subfigure}[t]{0.32\textwidth}
    \centering
    \includegraphics[width=\linewidth]{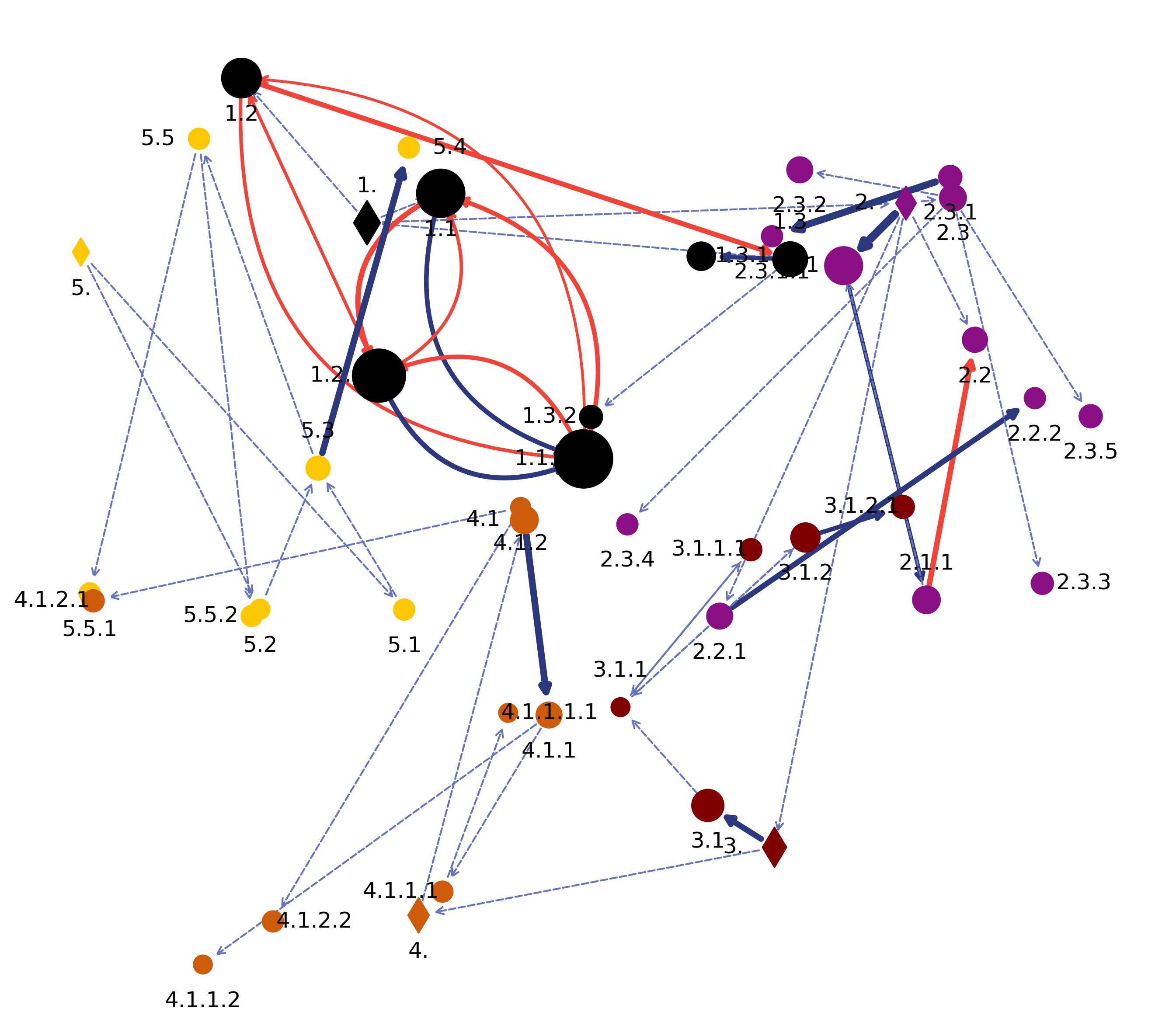}
    \caption{\small \textbf{Human FSM}: show higher variability in path selection and frequent revisiting of prior states. Larger node sizes indicate multiple re-engagements with key tasks, and sparse lateral connectivity suggests a tendency to follow one branch deeply rather than exhaustively traversing all parallel paths.}
    \label{fig:humanFSM}
  \end{subfigure}
  \hfill
  \begin{subfigure}[t]{0.32\textwidth}
    \centering
    \includegraphics[width=\linewidth]{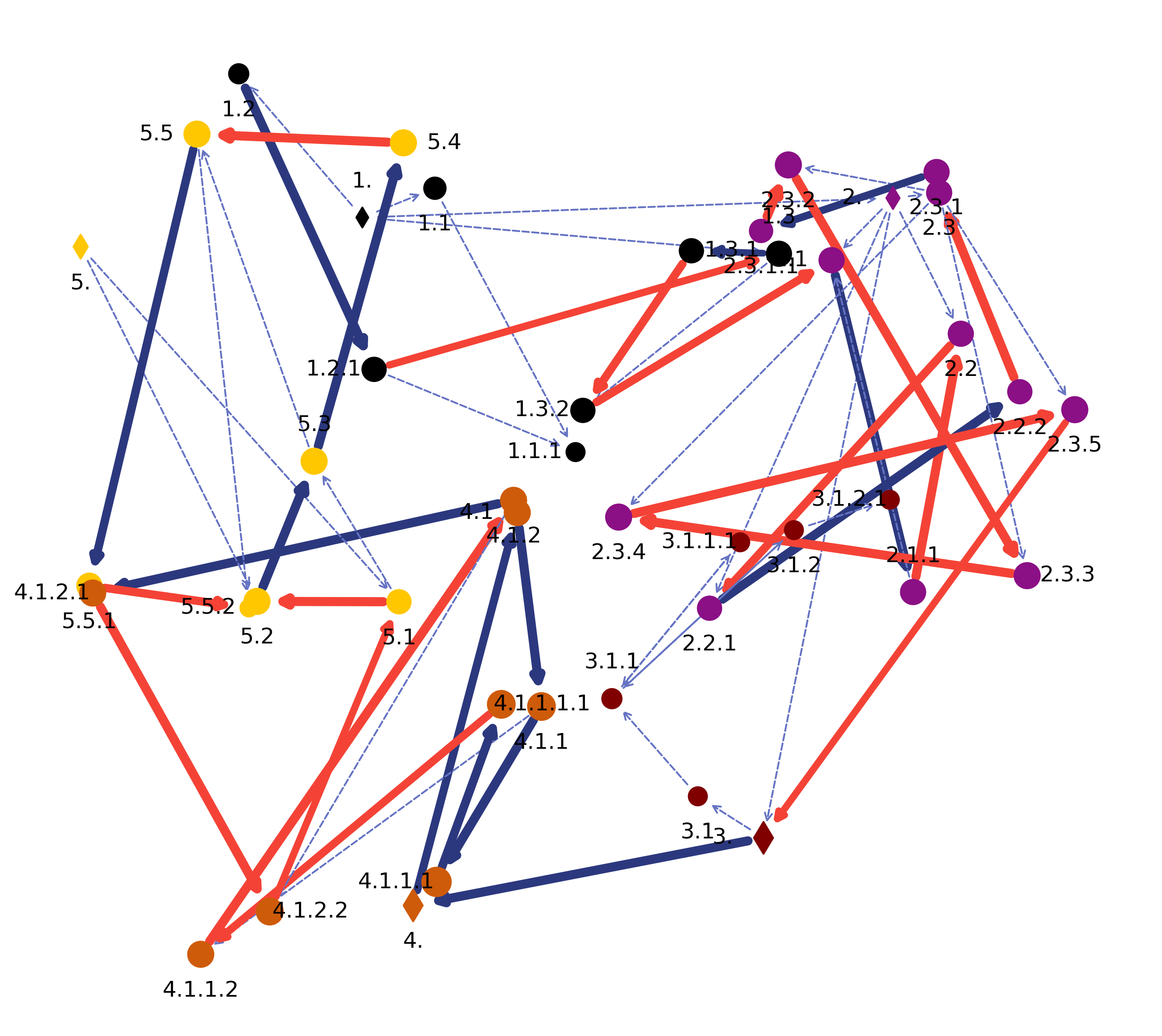}
    \caption{\small \textbf{O1 FSM}: reveals a highly exhaustive traversal style, covering nearly all nodes, including sibling tasks on the same hierarchical level. Because the task plan omits horizontal (same-level) transitions, O1’s breadth-first traversal across sibling nodes results in many red edges.}
    \label{fig:O1FSM}
  \end{subfigure}
  \hfill
  \begin{subfigure}[t]{0.32\textwidth}
    \centering
    \includegraphics[width=\linewidth]{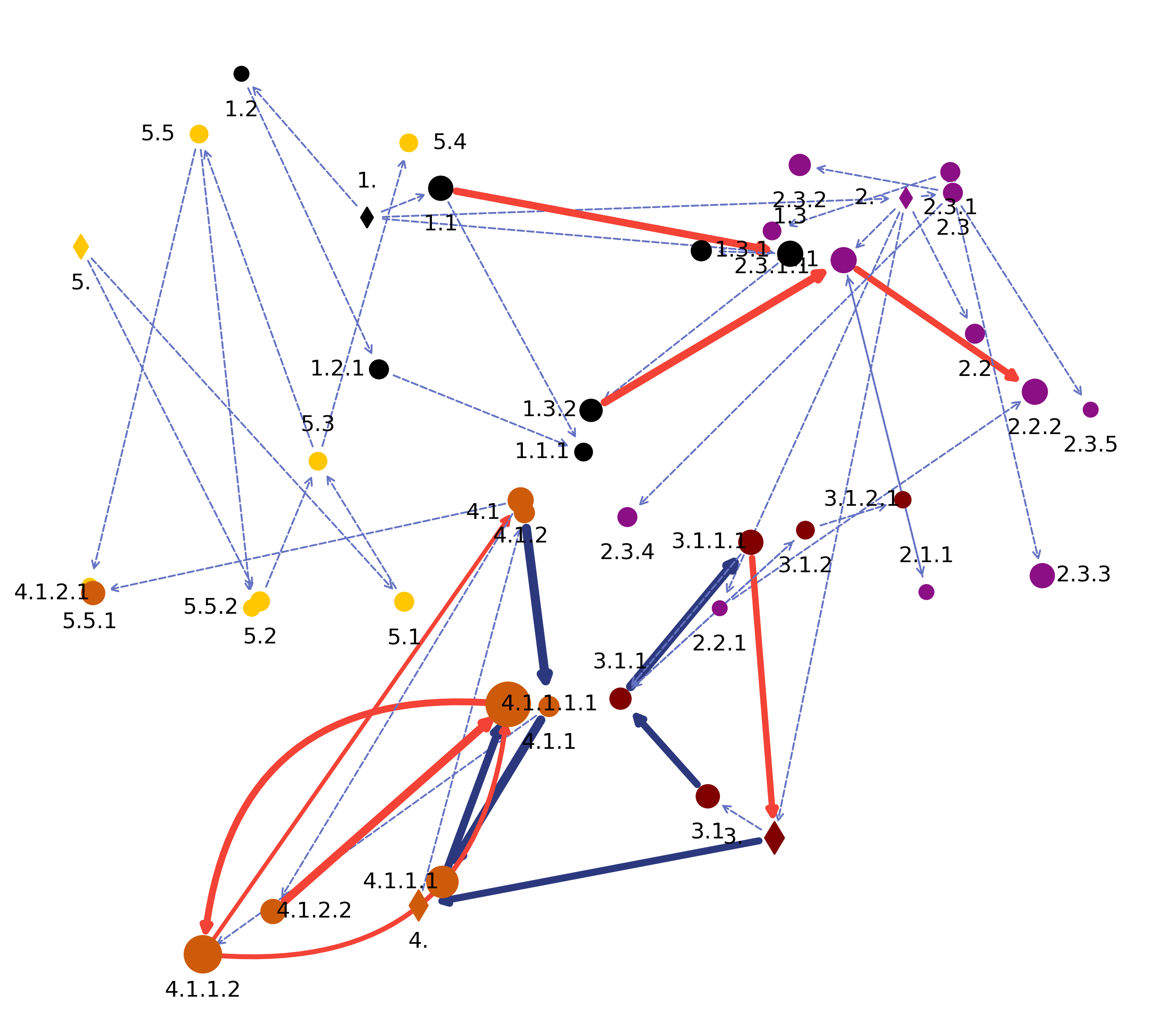}
    \caption{\small \textbf{R1 FSM}: exhibits a more restrained traversal than O1, engaging fewer sibling nodes and showing a transition pattern more consistent with depth-oriented exploration. The FSM displays wider coverage than human paths, but fewer red edges than O1, suggesting a hybrid strategy that partially aligns with human planning structure.}
    \label{fig:R1FSM}
  \end{subfigure}

\caption{\small \textbf{FSM visualizations} highlight differences in aggregate execution dynamics between humans and LLMs.  \textit{Node size} indicates frequency of revisits; \textit{edge thickness} reflects transition probability. \textit{Blue edges} denote planned transitions in task plan; \textit{Red edges} represent executed transitions not in the task plan. \textit{Solid edges} were executed; \textit{Dashed edges} were planned but not followed. We only display edges which have a threshold $>$ 0.5. Human workflows show selective, depth-oriented exploration; O1 follows an exhaustive, breadth-first pattern with many lateral transitions; R1 takes a more focused approach, partially resembling human strategies.}
  \label{fig:FSM_execution}
\end{figure}
\subsection {Can we model diverse legal workflows and flag unlikely steps?}
\label{sec:rq2}
Legal workflows are highly variable, shaped by context, judgment, and personal strategy. Even among trained law students, the same scenario can yield multiple valid paths. This diversity reflects the flexibility of legal reasoning, not noise. To support AI systems in this space, we analyze how workflows vary and introduce a lightweight ``workflow monitor'' tool to flag steps that diverge from common human patterns, prompting reflection without enforcing correctness.

\begin{figure}[htbp]
  \centering

  \begin{subfigure}[t]{0.24\textwidth}
    \centering
    \includegraphics[width=\linewidth]{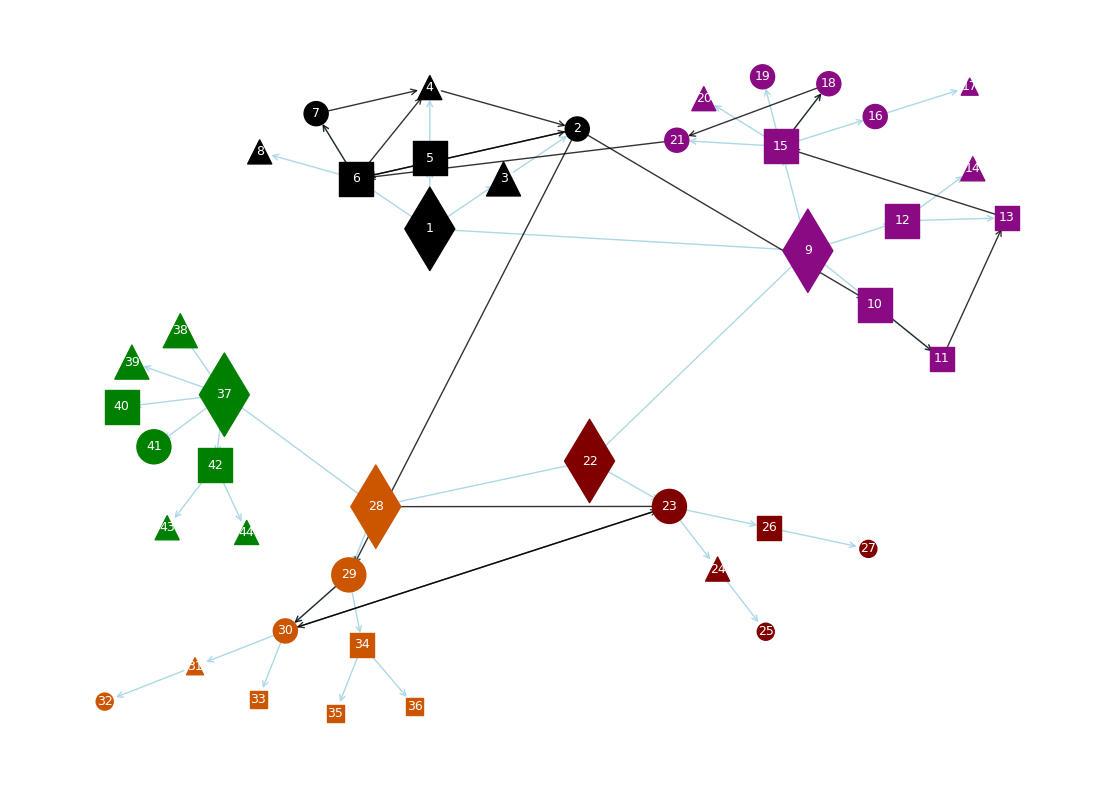}
    \caption{\small Law Student 1 executing ``Scenario 18''}
    \label{fig:image1}
  \end{subfigure}
  \hfill
  \begin{subfigure}[t]{0.24\textwidth}
    \centering
    \includegraphics[width=\linewidth]{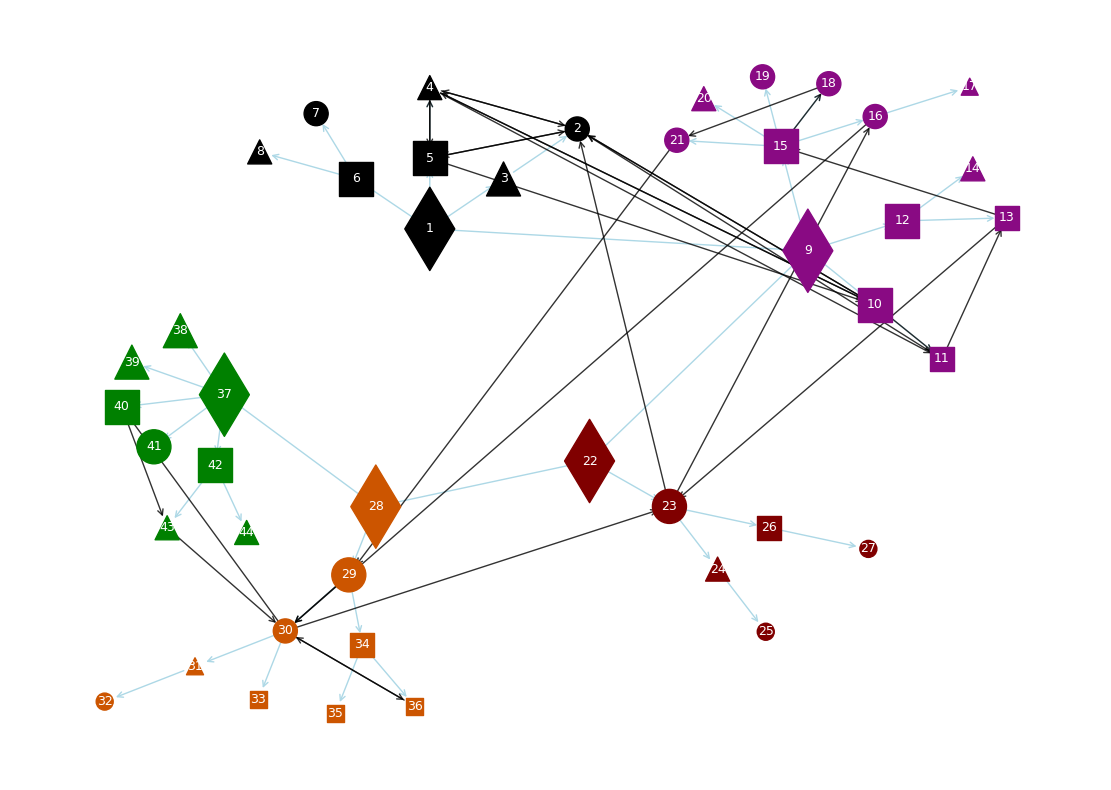}
    \caption{\small  Law Student 1 executing ``Scenario 10''}
    \label{fig:image2}
  \end{subfigure}
  \hfill
  \begin{subfigure}[t]{0.24\textwidth}
    \centering
    \includegraphics[width=\linewidth]{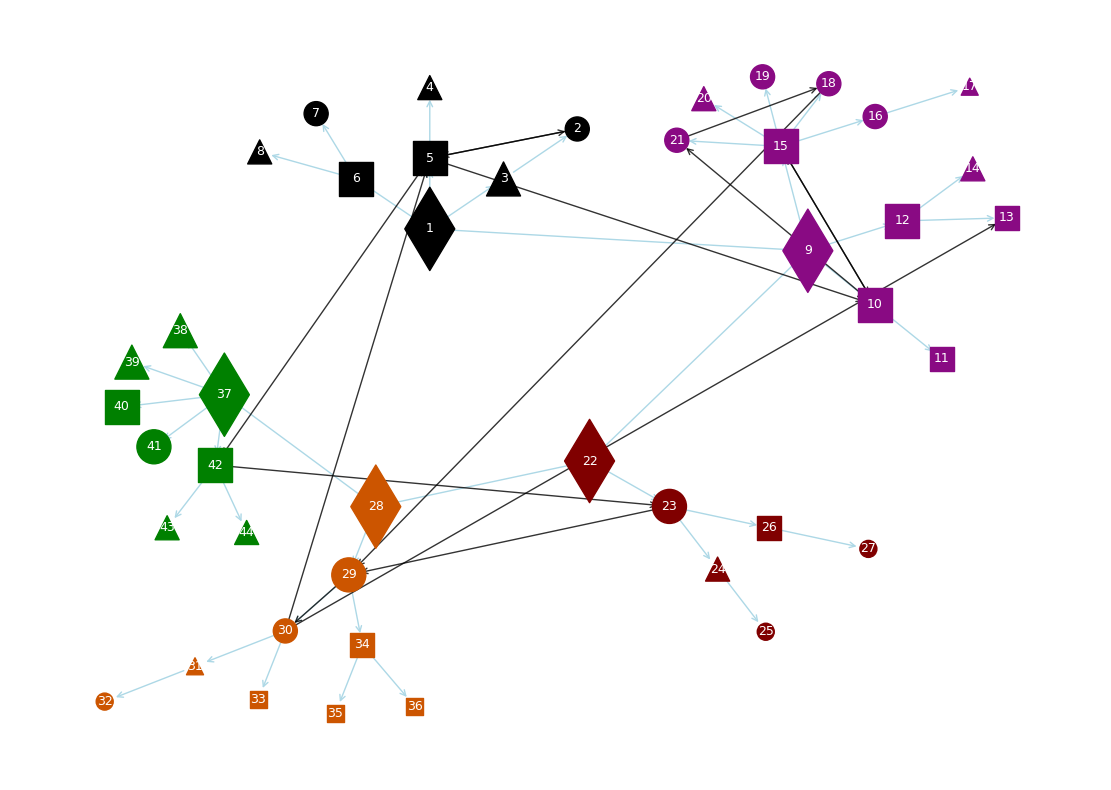}
    \caption{\small Law Student 1 executing ``Scenario 19 '' with 3 complexity parameters}
    \label{fig:image3}
  \end{subfigure}
  \hfill
  \begin{subfigure}[t]{0.24\textwidth}
    \centering
    \includegraphics[width=\linewidth]{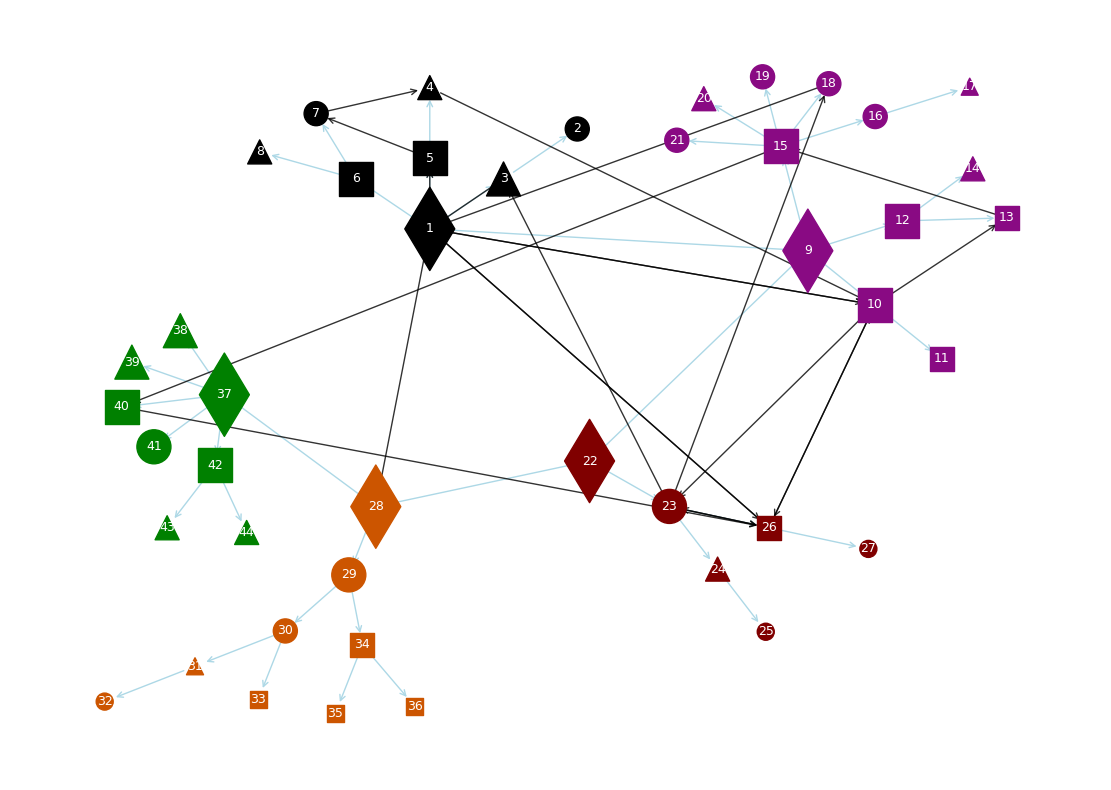}
    \caption{\small Law Student 2 executing ``Scenario 13.'' with 3 complexity parameters}
    \label{fig:image4}
  \end{subfigure}

  \caption{\small \textbf{All workflows matter}. Comparison of subtask connectivity across humans and tasks. The first two figures show the same human performing two different tasks. The last two figures show different humans performing tasks of similar complexity. Despite the variations, each workflow is a valid approach to completing the task. Descriptions of these scenarios are present in \ref{sec:client_elicitation}}
  \label{fig:four_images}
\end{figure}

\textbf{Characterizing Workflow Diversity.} We begin by analyzing workflows under two conditions: (1) the same law student executing different legal scenarios, and (2) different law students handling scenarios of similar complexity. Figure~\ref{fig:four_images} shows examples of subtask connectivity in these cases. The first two panels demonstrate how the same individual varies their strategy across contexts, while the latter two show \textit{different individuals approaching similar scenarios in distinct but valid ways}. These observations highlight the flexible, interpretive nature of legal work and highlight the limits of assuming a single canonical workflow. They also motivate the need for datasets and models that accommodate this variation in reasoning style.

To provide a more holistic view of these execution patterns across individuals and scenarios, we construct Finite State Machines (FSMs) based on observed reasoning paths (Figure~\ref{fig:FSM_execution}). FSM visualizations capture the underlying legal workflow language, showing typical state transitions, common loops, and pathway flexibility across scenarios.
By abstracting from individual differences, FSMs provide a holistic view of reasoning dynamics and reveal both normative paths and outlier transitions. This broader modeling enables lightweight monitors that can flag statistically atypical steps without enforcing a single correct sequence, encouraging reflection while respecting the interpretive flexibility inherent in legal work.

\begin{figure}[ht!]
\centering
\begin{subfigure}[t]{0.45\textwidth}
    \centering
    \includegraphics[width=\textwidth,clip, trim=0cm 1cm 8cm 0cm]{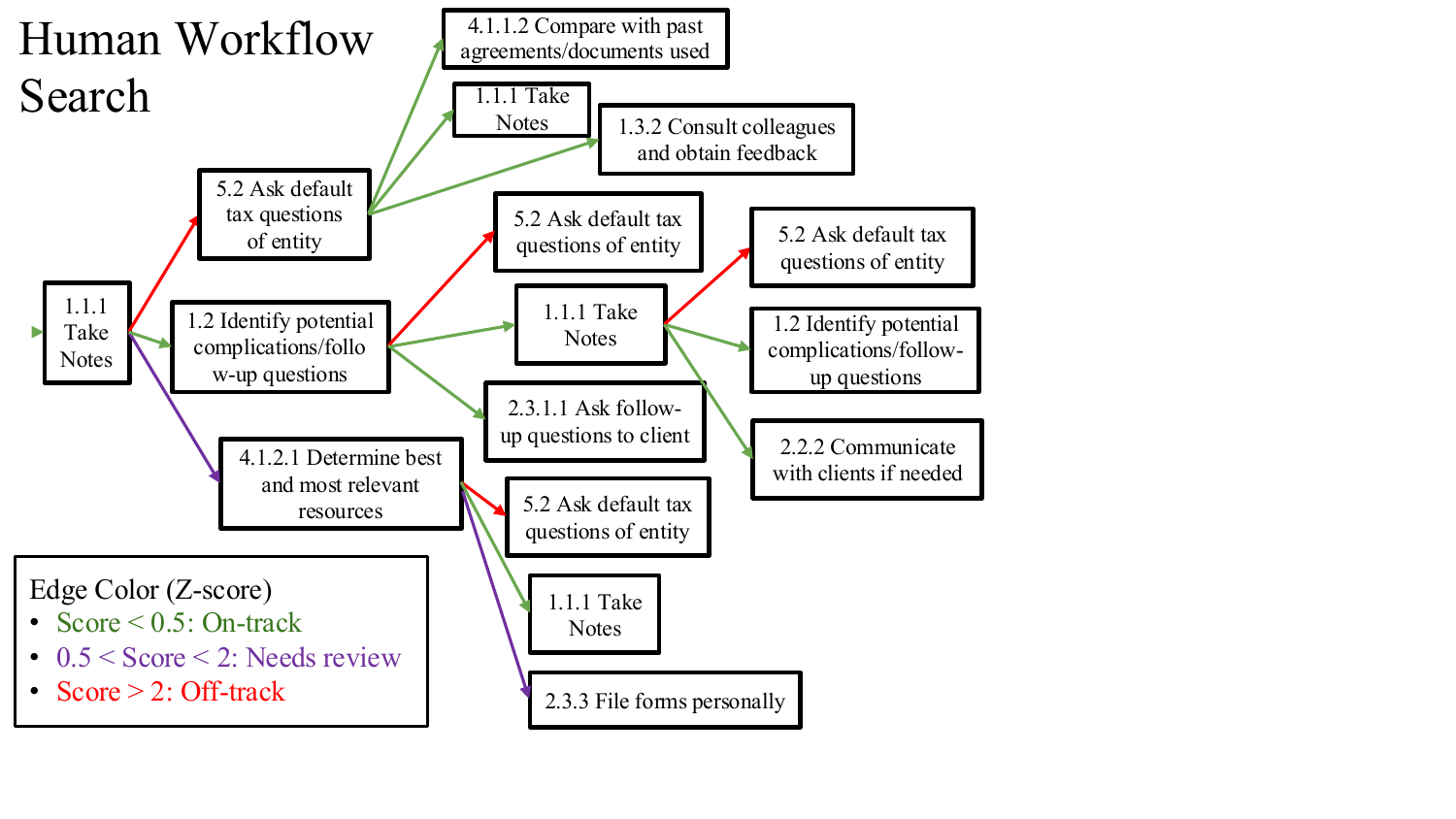} 
    \caption{\small Human workflows tends to favor commonly observed subtask sequences, such as note-taking followed by clarification and follow-up steps.}
    \label{fig:human-search}
\end{subfigure}
\hfill
\begin{subfigure}[t]{0.45\textwidth}
    \centering
    \includegraphics[width=\textwidth,clip, trim=0cm 1cm 7cm 0cm]{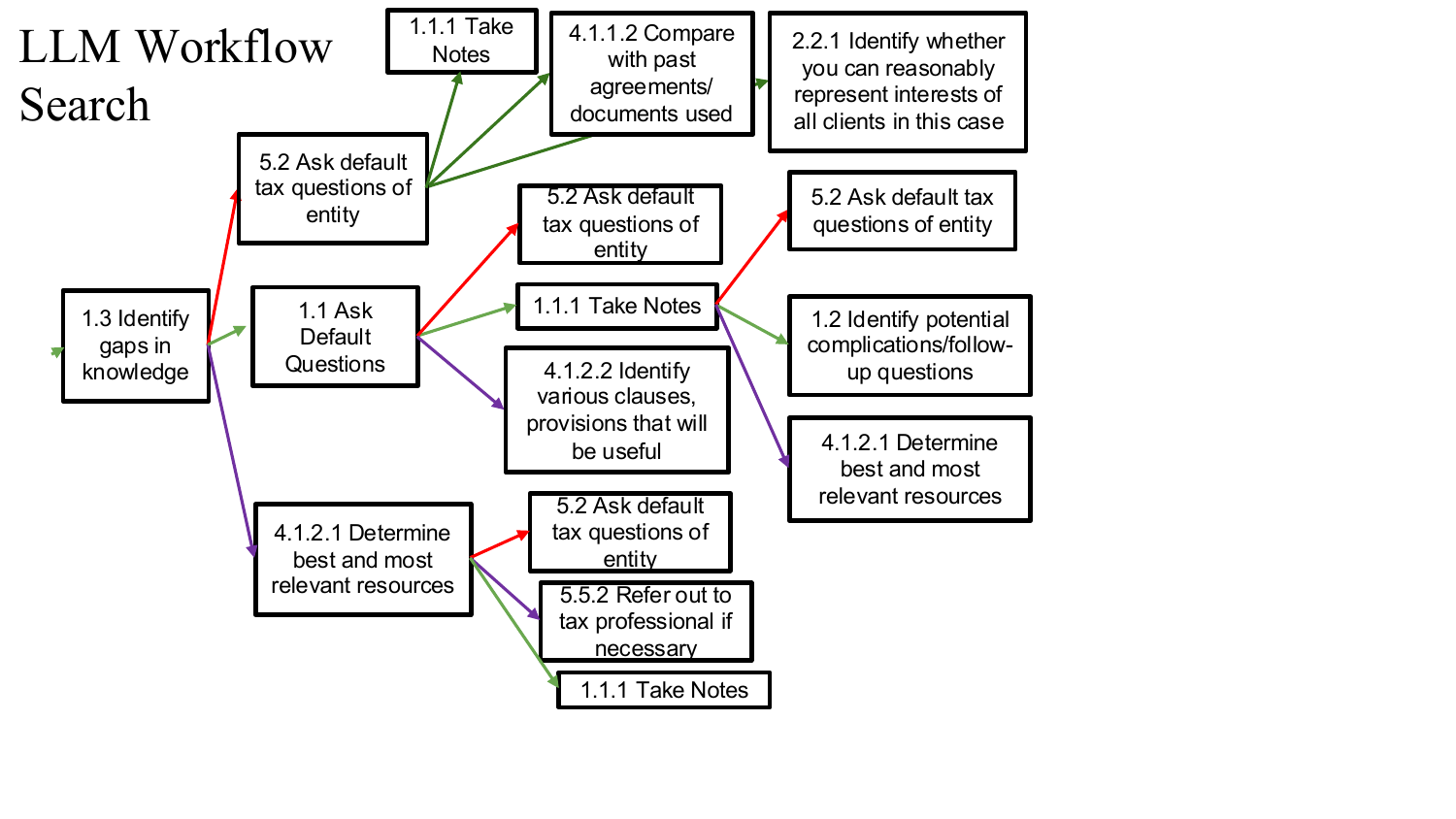}
    \caption{\small LLM workflows show jumping across task categories or introducing less context-sensitive steps, such as prematurely selecting resources or reverting to earlier-stage tasks.}
    \label{fig:LLM-search}
\end{subfigure}
\caption{\small \textbf{Step-level beam search visualizations} of Human and LLM workflows, colored by atypicality score (z-score) as predicted by a our workflow monitor : Transitions are scored by how well they align with human reasoning norms. Human workflows follow stable, commonly observed reasoning patterns, while LLM workflows exhibit greater variability and introduce less context-sensitive transitions.
} 
\label{fig:beam-search}
\end{figure}

\textbf{Modeling Atypicality.} To identify steps that diverge from typical human reasoning patterns, we train a step-level workflow monitor using the Llama-3.2-1B-Instruct model \citep{grattafiori2024llama3herdmodels} on eight human-generated workflows. At each step, the model predicts the most likely next action given prior context and assigns a perplexity score. We define a deviance score as the z-score of that step’s perplexity relative to the distribution observed in human training data. Lower scores indicate alignment; higher scores suggest atypicality, not as errors, but as moments that may benefit from review. Results (Appendix \ref{tab:deviance_band}) show that \textit{67.8\% LLM-generated steps align closely with human behavior}, while others fall into moderate and high deviation bands. These deviations often reflect context-insensitive decisions or uncommon execution paths. Rather than enforcing correctness, the workflow monitor provides a soft signal to support reflective execution. It helps surface unusual transitions in real-time. The beam search visualizations in Figure~\ref{fig:beam-search} show that human workflows tend to follow typical, well-aligned subtask sequences, while LLMs exhibit more scattered and atypical transitions.

\begin{table}[h]
    \centering
    \fontsize{7}{8}\selectfont 
    \renewcommand{\arraystretch}{1.2} 
    \setlength{\tabcolsep}{4pt} 

    \resizebox{\textwidth}{!}{ 
        \begin{tabular}{|p{4cm}|c|c|p{5cm}|}
            \hline
            \rowcolor{white}\textbf{Step Transition} & \textbf{Z-score} & \textbf{Flag (Deviation)} & \textbf{Interpretation} \\
            \hline
            2.3.5 Write memo to client $\rightarrow$ 3.1 Write memo & 0.11 & On track (0–0.5) & Aligned with expected workflow. \\
            \hline
            5.5.1 Advise on best entity form $\rightarrow$ 5.5.2 Refer to tax professional & 0.62 & Needs review (0.5–2) & Additional guidance may be needed. \\
            \hline
            4.1.2.2 Identify various clauses $\rightarrow$ 5.1 Ask default tax questions & 29.31 & Off track ($>$2) & Significantly deviates from human workflow; verify step. \\
            \hline
        \end{tabular}
    }
    \caption{\small \textbf{Z-scores} for each workflow step indicate alignment with human behavior. Steps are flagged as “on track,” “needs review,” or “off track,” highlighting whether they align with typical human reasoning, require further examination, or represent unusual transitions.}
    \label{tab:deviance}
\end{table}

  \subsection{What key decision points shape legal workflows, and how can AI assist at these junctures?}
\label{sec:rq3}

Some decisions in legal workflows disproportionately shape downstream outcomes. We refer to these as \textbf{meta-decision points}, points where the choice or action taken can reframe the task, influence later reasoning, or affect the final deliverable. Identifying these points helps us understand where workflow divergence is most likely and where AI assistance may offer the greatest value.
To surface meta-decision points, we combine graph-based analysis with human judgment. We compute two metrics across nodes in the task diagram: (1) \textit{Betweenness Centrality} – how structurally influential a node is in connecting the workflow and (2) \textit{Time Spent} – average human engagement time per node, signaling complexity or perceived importance
We also collect input from law students about which decisions they see as most critical. 

\begin{figure}[ht!]
\centering
\begin{subfigure}[t]{0.43\textwidth}
    \centering
    \includegraphics[width=\textwidth]{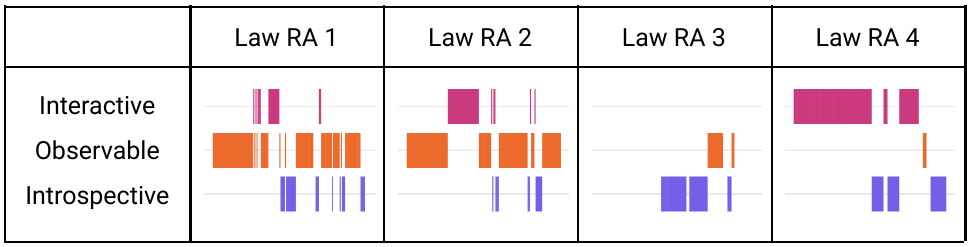} 
    \caption{\small Comparing the time spent on different cognitive tasks per RA}
    \label{fig:time_nodety}
\end{subfigure}
\hfill
\begin{subfigure}[t]{0.55\textwidth}
    \centering
    \includegraphics[width=\textwidth]{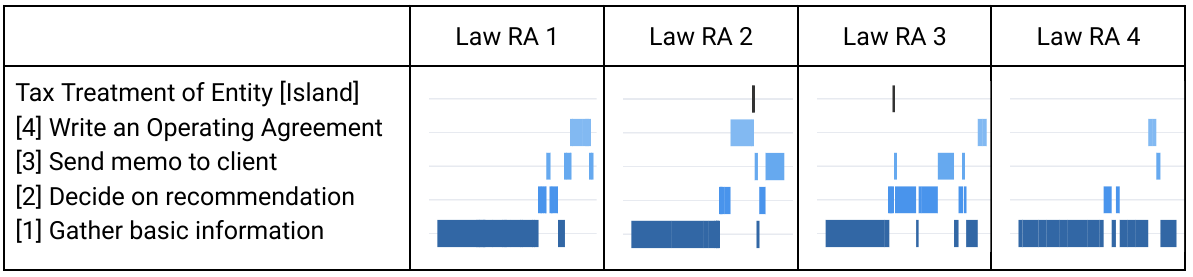}
    \caption{\small Comparing the time spent on different sub-tasks in the task plan per RA}
    \label{fig:time_subtask}
\end{subfigure}
\caption{\small \textbf{Comparison of activity timelines} grouped by sub-tasks and cognitive task types across all Law RAs} 
\label{fig:node_time}
\end{figure}

Results show that early-stage actions like note-taking and client questioning consistently rank high in centrality and time spent, especially in Subtask 1. Figure~\ref{fig:time_subtask} shows the timeline distribution grouped by Subtask. Subtask 1: Gathering basic information, which includes early-stage activities such as note taking and client questioning, occupies a significant portion of the overall timeline. Note-taking and Client questioning nodes consistently rank high in betweenness centrality, appearing in the top 3 in 70\% and 40\% of scenarios respectively, with neither appearing in just 10\% of the cases. 

Law students confirmed these as valuable moments for AI to assist in \textit{surfacing gaps and refining strategy}. In contrast, during later drafting (Subtask 4), students preferred \textit{AI in a review role}. Together, these signals help pinpoint where AI intervention can be both context-sensitive and high-leverage.

\begin{table}[t]
\centering
\scriptsize
\begin{tabular}{@{}>{\raggedright}p{0.09\textwidth}@{\hskip 1mm}p{0.24\textwidth}@{\hskip 2mm}p{0.21\textwidth}@{\hskip 2mm}p{0.19\textwidth}@{\hskip 2mm}p{0.22\textwidth}@{}}
\toprule
\textbf{Question} & \textbf{LLM Eval Trade-Offs} & \textbf{LLM (Eval) Rationale} & \textbf{Legal Expert (Eval)} & \textbf{Student (Writer) Defense} \\
\midrule
Question 1: Does the operating agreement \textbf{reflect all customer requirements}? & \textbf{Tension between formalism and practical norms}: LLM critiques the omission of specific operational details, reflecting a formalistic lens. However, both the expert and writer emphasize that excluding such details is not a flaw but a norm aligned with best practice. This highlights a trade-off between legal completeness and the flexibility favored in actual business structuring.
& Score: 3/5. Reasoning: The operating agreement encompasses broad operational and membership elements, but it \textcolor{green!50!black}{doesn't explicitly address all specific operational needs} ... 
& I don't agree with the LLM feedback which faults the operating agreement for just providing a governance framework without addressing specific operational details. But \textcolor{red}{those kinds of details would not generally be included in an operating agreement.}  
& No business consists of one document. There will be leases, use agreements etc. I do not see this as a flaw in the operating agreement. The \textcolor{red}{ more specific I make the operating agreement in terms of operations, the less flexibility there is and the more potential for conflict} concerning the operating agreement. Other documents can be more specific, but I think generality is good here. \\
\midrule
Question 2: Are there \textbf{clauses that might create unnecessary disputes} between stakeholders? & 
\textbf{Correctly flags potential risks, but underplays context}: LLM appropriately identifies concerns with missing valuation mechanisms and unanimity clauses, especially in dispute-prone situations. Yet, it may overemphasize these risks without accounting for context - the writer and expert view unanimity as a deliberate design choice suitable for a two-member LLC. This suggests a nuanced trade-off between structural safeguards and scalability.
& Score: 3/5. Reasoning: Several sections \textcolor{green!50!black}{might induce disputes, especially around transfer of interests, capital contributions, and the unanimity required for many decisions}. The \textcolor{green!50!black}{right of first refusal in Section 4.04(a) lacks detail on how the membership interest is valued} in conflicts, possibly leading to contention. Potential \textcolor{green!50!black}{deadlocks due to unanimity requirements without explicit dispute resolution mechanisms} for internal conflicts might also lead to issues
& The LLM expresses concern about the unanimity requirements.. But \textcolor{red}{with a 2 member LLC at the start, it's much less a problem}.. The \textcolor{green!50!black}{criticism of the valuation in the right of first refusal is absolutely accurate}, a real problem with the agreement. The \textcolor{green!50!black}{dispute mechanism resolution suggestion is worth considering}. With \textcolor{red}{2 members who have equal voting power, deadlock is unfortunately a built-in problem}...
& I agree that \textcolor{green!50!black}{requiring unanimity in decision making can become a problem if the company grows.} But it \textcolor{red}{can be an important mechanism for a small venture like this one}—you can’t have only two or three members who all disagree. It’s a potential for conflict, but it’s also an incentive for communication and compromise because you have no option to disregard your fellow members’ opinions.  ``There are no solutions, only trade offs.''\\
\midrule
Question 3: Are all \textbf{key terms clearly defined} in the document? 
& \textbf{Mixed accuracy in issue spotting}: LLM correctly identifies gaps in definition for some key terms (e.g., ``EBITDA multiplier''), though it misjudges others like ``capital contribution,'' which is defined. This reflects a trade-off in automated review : broad coverage and formal spotting versus precision and context sensitivity. 
& Score: 3/5. Reasoning: The document \textcolor{green!50!black}{lacks clarity on concepts like ``Capital Contribution,'' ``Pending Membership Interest,'' and ``EBITDA multiplier.}'' 
& \textcolor{red}{``Capital Contribution'' is defined}, but \textcolor{green!50!black}{EBITDA Multiplier and Pending Membership Interest are not}...
& \textcolor{red}{``capital contributions'' is defined} in Section 2... \textcolor{red}{``EBITDA multiplier'' and ``pending membership interest'' might be self-evident}... \\
\bottomrule
\end{tabular}
\caption{\small We compare \textbf{evaluations of student-drafted agreements} across five dimensions. The \textit{LLM (Eval) Rationale} column summarizes the LLM’s rubric-based evaluation of the agreement. The \textit{Legal Expert (Eval)} column reflects feedback on the LLM evaluation from a law professor, who is an academic with expertise in legal drafting pedagogy but limited real-world experience. The \textit{Law Student (Writer) Defense} column shows the student’s rationale for specific drafting choices. The \textcolor{red}{red} colored text indicates disagreement with the LLM while the \textcolor{green!50!black}{green} colored text indicates agreement. The \textit{LLM Eval Trade-Offs} column captures high-level patterns in how the LLM evaluates legal agreements. LLM evaluations tend to emphasize formal completeness, while human reviewers emphasize practical judgment and contextual appropriateness. This illustrates differing evaluation criteria between LLMs and domain-aware humans.}

\label{tab:rubric_eval}
\end{table}
\subsection {Interpreting Quality: Divergence in Evaluating Agreements}
\label{sec:eval_quality}
While structural comparisons help illuminate the \textit{how} of legal reasoning, understanding the \textit{why} behind legal choices requires examining how legal workflows are evaluated. Rubric-based LLM assessment frameworks emphasize formal completeness and surface-level consistency. However, these criteria often diverge from the practical, context-sensitive reasoning employed by legal experts. To explore this disconnect, we compared LLM evaluations of human-authored operating agreements with critiques from legal experts and reflections from the student authors themselves.

As shown in Table~\ref{tab:rubric_eval}, LLMs frequently flagged omissions that legal professionals defended as intentional, context-appropriate decisions. For instance, critiques about missing operational specifics overlooked established norms; both the expert and the writer emphasized that such details are best handled in ancillary documents rather than in the operating agreement itself. Similarly, concerns about unanimity requirements failed to take into account the strategic value of consensus in small, closely held businesses, where unanimity can encourage communication and mitigate power asymmetries.. These examples reflect a broader pattern: LLMs often apply generalized standards without recognizing domain-specific practices, client intentions, or the trade-offs that shape real-world legal drafting.

That said, LLM evaluations are not without merit. They surface valuable concerns that human drafters may overlook. In one example, the LLM correctly identified a lack of valuation detail in the right-of-first-refusal clause, an omission the legal expert agreed could generate future disputes. It also flagged the absence of explicit dispute resolution provisions, prompting both expert and student reconsideration. These instances demonstrate how LLMs, with their exhaustive and systematic lens, can serve as a valuable second-pass reviewer, catching latent issues even when their evaluative logic is imperfect.
Expert assessments also revealed that LLM-generated agreements are not always inferior. Out of the two agreements that the expert evaluated, they judged one of the LLM-drafted agreement to be \textit{preferable} to the student version. The LLM’s draft was praised for its greater clarity, completeness, and improved treatment of stakeholder roles, particularly in how it incorporated provisions for children, which the student agreements neglected. While some of the LLM’s choices (e.g., voting thresholds) were seen as potentially confusing, the agreement overall was deemed more thoughtfully constructed. This comparison shows that LLMs are capable not only of critique but also of producing high-quality legal drafts, sometimes outperforming novice human writers.

Ultimately, these comparisons highlight that LLMs currently adopt a formalist perspective, privileging structure, coverage, and textual clarity over the more nuanced, judgment-driven reasoning that human experts rely on. Supporting legal professionals in high-stakes contexts will require AI systems to incorporate better human-centered heuristics, striking a balance between rigor and adaptability, and structure and pragmatism. Bridging this gap requires not only more capable models but also training data that captures how real practitioners navigate ambiguity, trade-offs, and evolving client needs—precisely the kind of insight \texttt{LawFlow} is designed to provide.


\begin{figure}[h!]
  \centering
  \vspace{-2mm}
  \includegraphics[width=0.99\columnwidth,clip, trim=0cm 5.5cm 6.5cm 0cm]{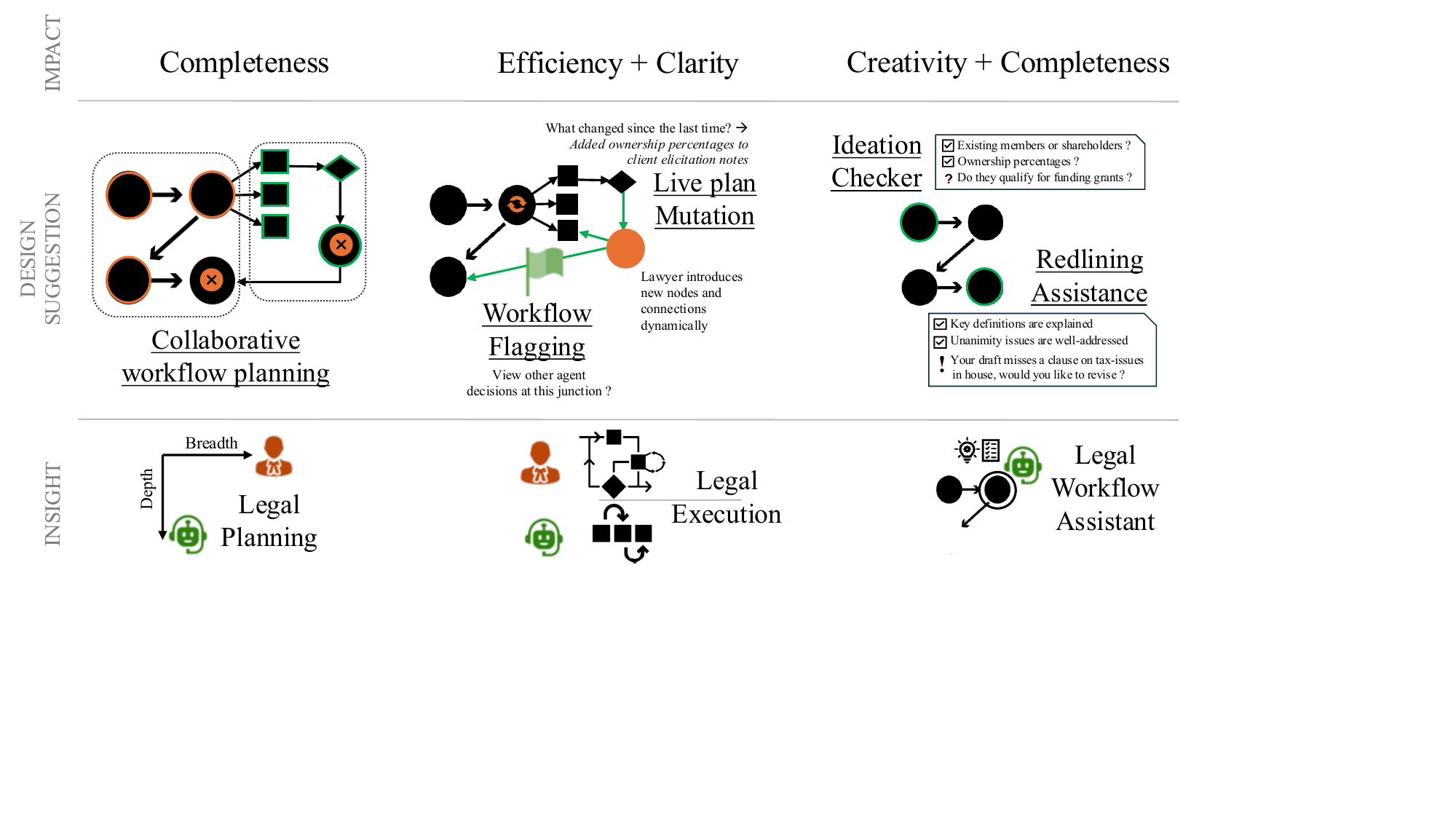}\vspace{-2mm}
  \caption{\small \textbf{Design suggestions} based on insights learned from \texttt{LawFlow} include collaborative workflow planning, workflow flagging and AI assistance with specialized roles which can aid human goals of clarity, efficiency, creativity and completeness.}
  \label{fig:design}
\end{figure}

\section{From Insights to Actions: Design Principles for Collaborative Legal Assistants}

Building directly on the empirical findings in Section \ref{sec:sec4}, we outline a next-generation legal assistance framework that treats AI not as a replacement for lawyers or judges, but as a collaborative partner whose rigor complements human nuance. A key insight from \texttt{LawFlow} is that human and AI reasoning and planning often differ in structure, adaptability, and rigor. These differences can be harnessed to improve outcomes rather than being viewed solely as shortcomings. Each design pillar below maps to a concrete observation from our experiments and sketches how legal practitioners and AI systems can collaborate more effectively.

Currently, workflow quality is primarily judged by evaluating the final output, which is the agreement. While this focus on final document quality reflects real-world expectations, it offers a limited view of the broader reasoning and planning process. To capture workflow quality more holistically, we propose supplementing this output-based evaluation with process and behavior-oriented metrics that better aid human goals of clarity, efficiency, creativity, and completeness.

Specifically, we propose three metrics:
\begin{enumerate}[noitemsep, topsep=0pt, left=1em] 
    \item \textbf{Decision consistency}, which measures how reliably participants revisit or stick with their earlier choices helping promote both completeness and clarity across the workflow.
    \item \textbf{Workflow completeness}, which evaluates whether key subtasks are addressed in a logical and thorough sequence, allowing for greater task completeness and process rigor.
    \item \textbf{Evaluation of point performance}, which focuses on high-impact decision moments that disproportionately affect downstream outcomes. Careful assessment and assistance at these points improve not only task completeness and clarity but also aid creative problem-solving, as participants are encouraged to generate well-justified solutions rather than relying on rote templates. For instance, selecting the appropriate business entity structure (e.g., an LLC vs. a C-corp) is a pivotal choice that affects liability, taxation, and document organization. An accurate and well-supported decision at this junction enhances clarity throughout the agreement, improves drafting efficiency by reducing revisions, and enables creative structuring options that better align with client goals. By emphasizing point performance alongside overall process quality, we can better assess where strong legal judgment, rather than just procedural diligence, drives superior results.
\end{enumerate}

We outline design principles (illustrated in Figure \ref{fig:design}) aligned to these metrics, each aimed at improving legal workflows through human-AI collaboration.

\paragraph{1. Collaborative Planning – Human Depth + AI Breadth}
We observe that humans build deep, modular plans, while LLMs generate exhaustive but rigid checklists. Rather than judging one style as superior, we can merge their strengths.
\begin{itemize}[leftmargin=1.5em,noitemsep]
\item \textit{Human skeleton.}  The lawyer sketches 3–6 top‑level goals, mirroring the deep, modular structures observed in human plans.
\item \textit{AI breadth pass.}  AI expands each goal into a collapsible checklist, surfacing edge cases the lawyer may overlook, leveraging the AI's exhaustive but flat planning style.
\item \textit{Reconciliation view.}  A side‑by‑side diff lets the user prune or promote AI branches, yielding a jointly developed hybrid plan.
\end{itemize}
This intuition of mutual support aligns with lessons from judicial decision-making research \citep{spamann2016justice, spamann2024can}, which shows that while human judges often deviate from strict legal precedent due to extralegal factors, such as sympathy, LLMs exhibit a more rigid, formalist approach similar to that of law students. Whether formalism is an asset or a liability depends on the task at hand. In drafting and compliance, a systematic approach can help reduce overlooked details, but in complex, high-stakes judgments, the human judge’s nuanced weighing of facts and equities may be indispensable. This hybrid design could enhance both \textbf{workflow completeness} and \textbf{evaluation of point performance} by encouraging lawyers to combine structured human reasoning with AI-surfaced contingencies. It directly supports the human goal of task completeness, helping lawyers structure more thorough and resilient workflows without sacrificing focus.

\paragraph{2. Adaptive Execution Engine \& Reflective Monitoring}
During execution, law students routinely \emph{loop back} and reorder tasks, whereas LLMs predominantly execute through a plan once, in strict sequence. To support human adaptability without sacrificing coherence, we propose:

\begin{enumerate}[leftmargin=1.5em,noitemsep]
\item \textit{Live plan mutation.}  When new facts emerge, users can drag and rearrange tasks; the LLM automatically rethreads downstream dependencies, maintaining coherence while allowing dynamic adjustment.

\item \textit{Workflow‑diversity monitor.}  A lightweight model flags statistically unusual moves relative to prior human traces, prompting reflective checks without enforcing strict conformity.

\end{enumerate}
These tools could streamline plan adjustments and spotlight deviations, promoting greater efficiency and clarity. By supporting \textbf{decision consistency} and \textbf{workflow completeness}, they help maintain coherence while still allowing creative flexibility. Reflection could also be deepened through lightweight nudges from LLMs, such as surfacing stale assumptions or offering comparative junction-specific peer views. 

\paragraph{3. Meta‑Decision Point Assistance}
In our work, betweenness centrality and time-spent analyses identify note-taking, client-elicitation questions, entity selection, and reviewing operating agreements as high-leverage nodes that impact the workflow of the drafting process. Thus, we propose design suggestions that modify the LLM’s role at these meta-decision points. 
\begin{enumerate}[leftmargin=1.5em,noitemsep]
\item \textit{Front‑loaded prompts.}  During client intake, the assistant auto‑suggests clarifying questions and surfaces similar fact patterns, ensuring critical details are captured before substantive work begins.

\item \textit{Review‑only drafting.}  In later stages, the LLM defaults to ``redline \& comment'' mode, offering critiques and consistency checks unless the user explicitly requests auto‑rewrites.
\end{enumerate}
Thus, AI effort is focused exactly where it can significantly aid the lawyer, and the system could enhance human creativity (by encouraging thoughtful exploration of alternatives) and completeness (by ensuring pivotal issues are surfaced early and handled carefully). This strategy directly strengthens \textbf{point performance}, ensuring workflows are anchored by strategically sound choices.

Beyond the immediate design suggestions, the \texttt{LawFlow} dataset also can be synthetically extended into multiple ``near‑miss'' variants by reordering steps or omitting low‑value actions. These chain-of-thought (COT) style expansions would multiply the dataset while preserving a realistic, student‑level voice, enabling richer modeling of developmental reasoning. Such extensions open the door to broader outcomes, including peer-mentoring systems, scaffolded checklists, and alternative-path exemplars. Together, these developments could create a living library of legal workflows, helping users not only complete tasks more effectively but also continuously strengthen their reasoning and drafting strategies in ways grounded in authentic legal practice.

In this broader collaborative framework, AI-supported tools built around \texttt{LawFlow}-style metrics do more than promote formal completeness and decision consistency, they accelerate drafting workflows, improve document clarity and robustness, and spark creativity by surfacing overlooked opportunities and trade-offs. Rather than replacing human expertise, such systems scaffold better professional judgment, helping legal practitioners produce work that is both technically sound and strategically optimized.

\section {Conclusion}

This paper presents \texttt{LawFlow}, a dataset designed to capture the complete decision-making workflows of trained law students performing realistic legal drafting tasks. Unlike prior datasets that focus on isolated legal subtasks, \texttt{LawFlow} models the process of legal reasoning, revealing how humans adapt their plans based on uncertainty, incomplete information, and evolving context. 

Our findings are grounded in two complementary methodologies: unobtrusive logging of human role‑play sessions and a structured ``next‑step’’ prompting setup for LLMs. While this design allows direct comparisons, it also imposes limitations. The linearity observed in LLM workflows may partially reflect prompting artifacts rather than purely model behavior. Black-box LLM inference implies we capture only externalized action sequences, not internal reasoning. Future work should replicate these findings using alternative prompting schemes or think-aloud protocols to better align legal experts and AI comparisons. Nevertheless, aligning the two views on the same underlying client scenario and background documents makes it possible to identify systematic points of divergence, or places where an AI monitor could add value by signaling context-insensitive moves or by recommending clarifying questions.
Taken together, \texttt{LawFlow} serves as:
\begin{enumerate}[leftmargin=1.5em,noitemsep]
    \item a \textbf{divergence lens} for examining how legal reasoning pathways vary across agents
    \item a \textbf{test bed} for lightweight workflow monitors that flag atypical steps without prescribing a single “correct” path
    \item a \textbf{seed corpus} for designing collaborative assistants that visualise progress, suggest alternatives, and intervene at high‑leverage decision nodes.
\end{enumerate}

Beyond these empirical contributions, we propose design principles for next-generation legal AI systems, rooted in our observations: hybrid planning that combines human depth with AI breadth, adaptive execution support to maintain consistency and flexibility, and targeted assistance at pivotal decision-making moments to foster clarity, creativity, and completeness. Extending \texttt{LawFlow} with additional traces, richer baseline models, and more diverse legal scenarios will deepen these insights. Even in its current form, \texttt{LawFlow} demonstrates that modeling the decision process, rather than isolated tasks, opens up promising new directions for building AI systems that truly complement the iterative and context-sensitive nature of real-world legal practice.
\section{Limitations}

Creating the \texttt{LawFlow} dataset involved several challenges. First, capturing end-to-end legal workflows is inherently difficult as such datasets are virtually nonexistent, likely due to the high human effort, expert oversight, and coordination required. Our approach relied on a structured human task plan to guide data collection. While this kind of framework was necessary for consistent annotation and workflow tracing, it also introduced a potential artifact: by providing an explicit structure, it subtly influenced participants to follow that structure, potentially reducing the natural variability and nonlinearity typical of real-world legal practice. Second, the entity formation scenarios are intentionally simple (Appendix \ref{sec:client_elicitation}), which helps isolate reasoning patterns but may underrepresent the complexity of real-world legal practice. Future versions will incorporate more nuanced, expert-verified cases.

Third, modeling authentic legal behavior required accommodating the unpredictability of human actions. Participants often deviated from the intended tool use, for example, by consulting external resources like Westlaw or Google, which, while reflective of real legal workflows, introduced noise into our structured logging. Rather than treat these deviations as errors, we view them as important signals about how legal practitioners actually behave under realistic conditions. Future iterations of LawFlow may seek to explicitly log or integrate such external tool usage to better capture the full spectrum of legal reasoning.

Finally, both annotation and evaluation introduce additional complexity. Standard metrics like task time are confounded by learning effects, as law student annotators become more efficient over repeated scenarios. To address this, we advocate for behavior-based measures such as decision consistency to better track evolving expertise. Similarly, disagreement between human and AI outputs is expected, given the inherent subjectivity of legal reasoning. Rather than treating these divergences as failures, we treat them as valuable points for exploring human-AI alignment and understanding where automated systems fall short of expert judgment.



\section*{Acknowledgments}
This work was mainly supported by the Agent Benchmark Grant from the Open Philanthropy. We also thank Minnesota NLP group members for providing us with valuable feedback and comments on the initial draft.
We are grateful to Kiri Somermeyer at UMN Law's Business Law and Entrepreneurship Clinic for sharing business formation scenarios and offering valuable feedback.


\bibliography{colm2025_conference}
\bibliographystyle{colm2025_conference}

\appendix
\section{Appendix}


\subsection{More about Data Collection}
\label{sec:data_sec}

\subsubsection{More about the expert-informed task plan}
\begin{figure}[h!]
  \centering
  \vspace{-3.5mm}
  \includegraphics[width=0.70\columnwidth,clip, trim=0cm 0cm 0cm 0cm]{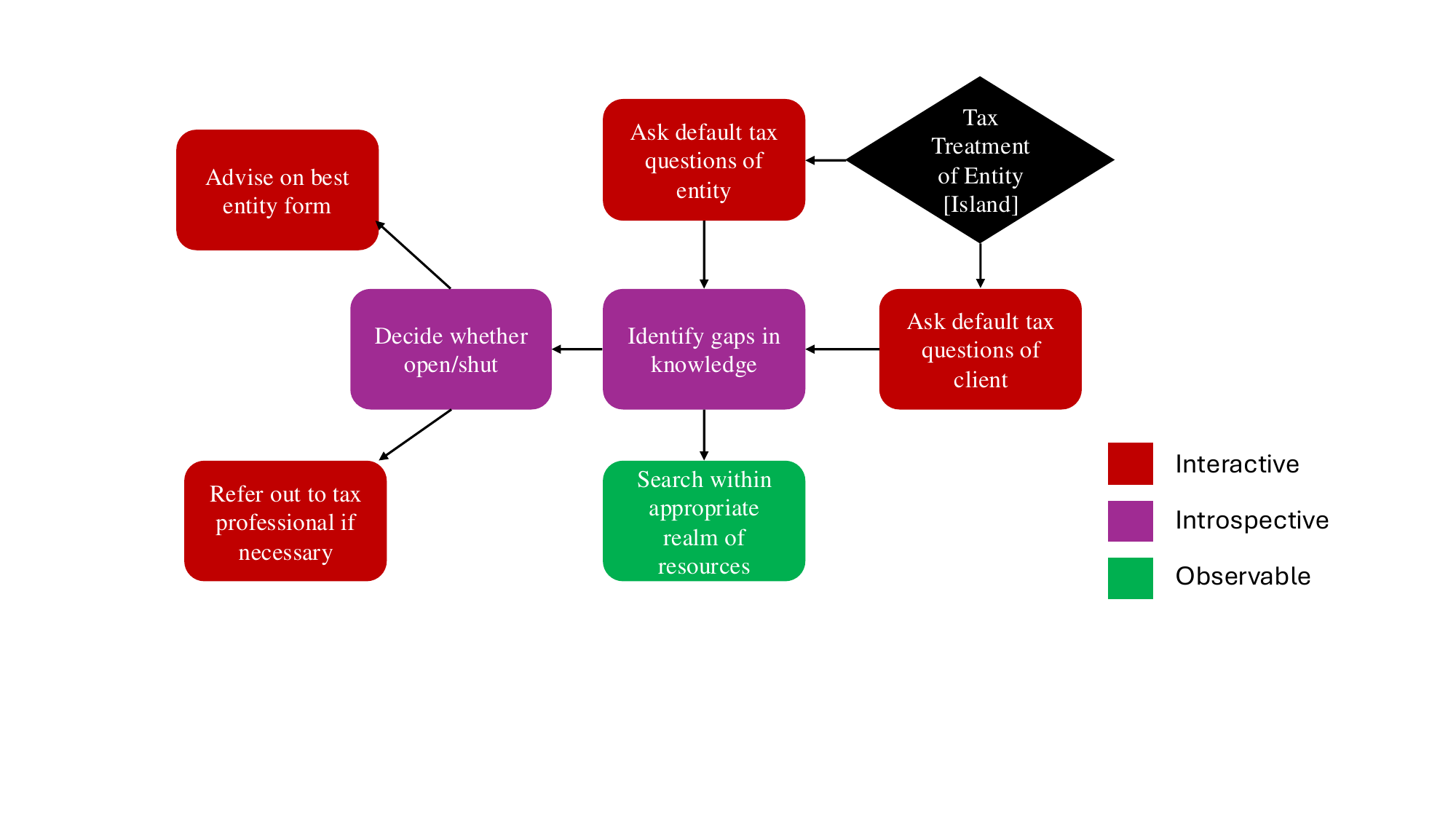}
  \caption{\small \textbf{Subtasks involved in dealing with Tax considerations} (considered mostly optional in our scenarios)}
  \label{fig:memo}
\end{figure}

We conducted interviews with senior law faculty and students to model the end-to-end workflow of business entity formation leading to an Operating Agreement. The expert-informed task plan in Figure \ref{fig:task_diagram} spans three levels: major workflow stages (inter-subtask), steps within each stage (intra-subtask), and individual actions tagged by cognitive modality - introspective, interactive, and observable. Key subtasks include gathering client information, forming legal recommendations, drafting memos, composing Operating Agreements, and optionally assessing tax implications (details of this are provided in Figure \ref{fig:memo}). 

\subsubsection{Scenarios informing Data Collection}
\label{sec:client_elicitation}
Here, we present some of the entity formation scenario's we have mentioned in our analysis in this paper. In addition to subtask tagging, each scenario is annotated with two metadata sets: \textit{complexity} and \textit{nuance} parameters, which are described in Table \ref{tab:complexity-nuance}. These annotations allow us to capture factors that influence the legal task's overall structure and difficulty and the role-play's interactive dynamics. \textit{Complexity parameters} identify structural features of a scenario that are likely to impact the scope and duration of the legal work and the outputs produced (e.g., operating agreement, client memo). These parameters influence how the lawyer navigates the workflow and which subtasks are emphasized, revisited, or escalated. \textit{Nuance parameters}, by contrast, shape the interactivity and dynamism of the roleplay itself. These factors make the simulation more realistic and help model the types of human-centered challenges lawyers regularly face. By systematically varying these parameters across scenarios, we are able to study how different contextual and interpersonal factors shape legal task execution. This, in turn, supports a more robust analysis of legal workflows, the adaptability of experts, and the design of systems that can assist with or simulate professional legal reasoning in diverse settings. Table \ref{tab:scenario-diversity} summarizes the diversity and complexity of the scenarios in our dataset. Table \ref{tab:business-entity-metric} provides statistics on the types of business entities covered in our dataset and the lengths of the outputs and tool usage associated with each.  

\begin{table}[ht]
\scriptsize
\centering
\renewcommand{\arraystretch}{1.2}
\begin{tabularx}{\textwidth}{X|X}
\toprule
\textbf{Complexity Parameters} & \textbf{Nuance Parameters} \\
\midrule
\begin{itemize}[leftmargin=*]
    \item Whether a client memo is needed
    \item Whether separate lawyers are required for conflict management
    \item Whether routing to a different lawyer is necessary due to lack of expertise
    \item Number of individuals involved in the business
    \item Nature of the business (e.g., regulated industry, high-risk sectors)
    \item Business expansion plans
    \item Number of investment sources
    \item Diverging interests among founders or stakeholders
\end{itemize}
&
\begin{itemize}[leftmargin=*]
    \item Need for follow-up with the client after the initial information-gathering session
    \item Need for follow-up after delivering the first version of the operating agreement
    \item Whether the client is overprepared (e.g., brings documents, has done prior research)
    \item Whether the client is underprepared (e.g., lacks familiarity with legal or business concepts)
    \item Whether multiple client sessions are needed to arrive at an entity choice decision (e.g., due to complex funding structures or high-stakes investments)
\end{itemize}
\\
\bottomrule
\end{tabularx}
\caption{\small Parameters for \textbf{Complexity and Nuance} in Legal Consultations}
\label{tab:complexity-nuance}
\end{table}

\begin{figure}[ht!]
\centering
\begin{subfigure}[t]{0.55\textwidth}
    \centering
    \includegraphics[width=\textwidth,clip, trim=2cm 0cm 2cm 0cm]{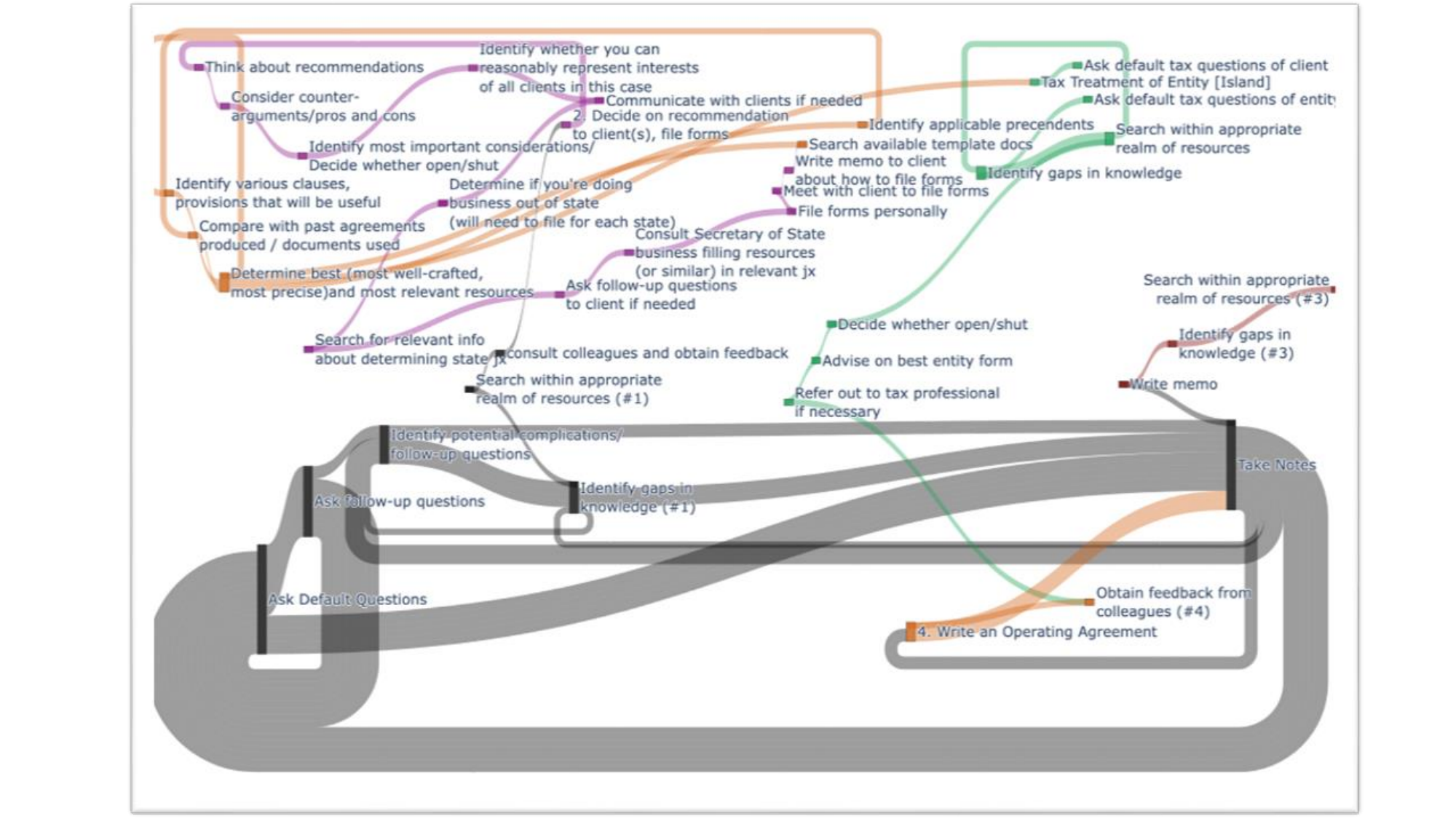} 
    \caption{\small \textbf{Human execution workflow for Scenario 1}: A and B are avid anglers aiming to promote fishing in our area by hosting free educational workshops for children and adults. They plan to form a business to manage gear purchases, raffles, and possible future funding, but want to avoid
complex formalities and do not intend to make a profit. They are interested in understanding what kind of startup agreement they need. }
    \label{fig:sankey1}
\end{subfigure}
\hfill
\begin{subfigure}[t]{0.43\textwidth}
    \centering
    \includegraphics[width=\textwidth,clip, trim=1cm 0cm 0cm 0cm]{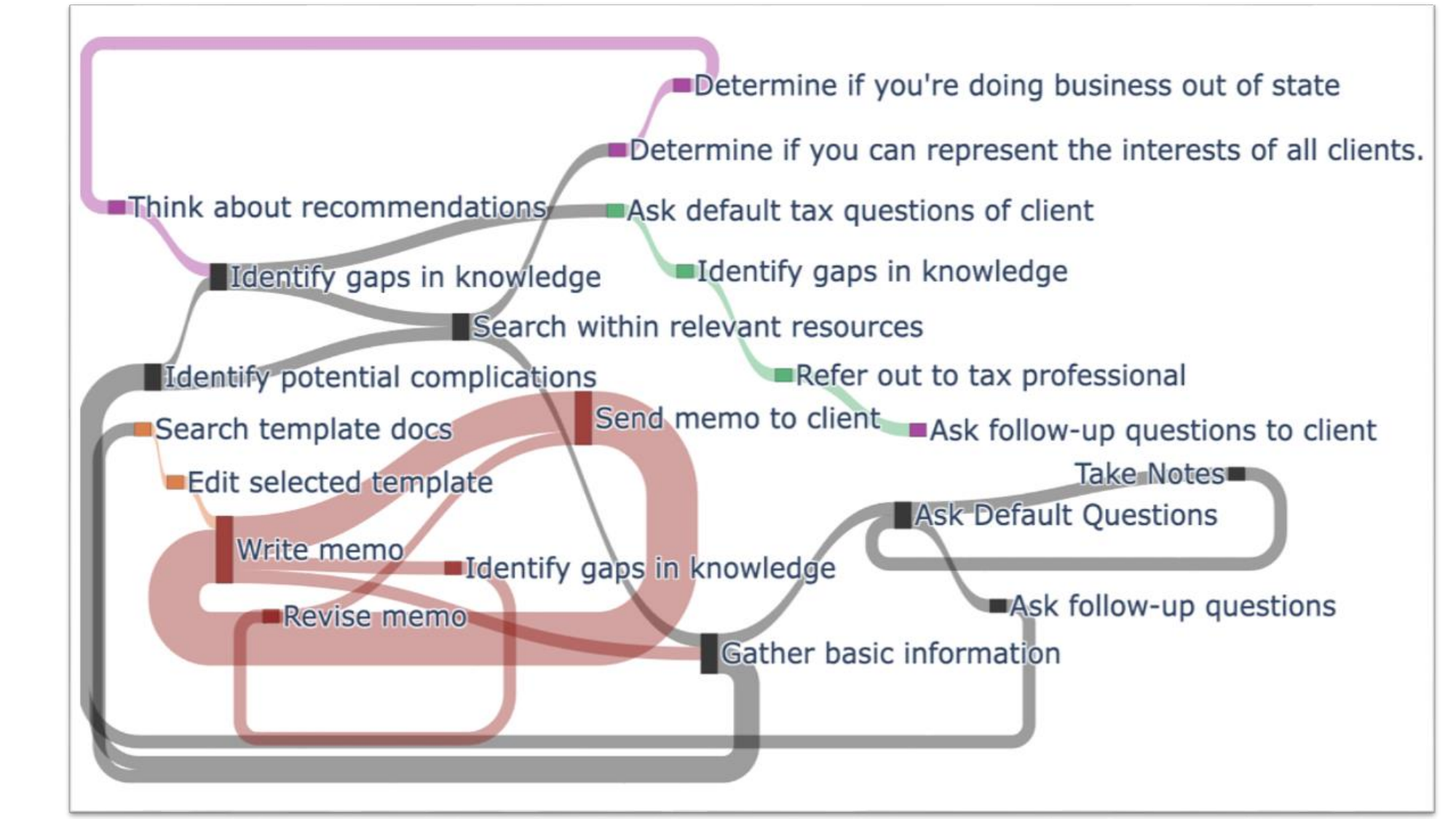}
    \caption{\small \textbf{Human execution workflow for Scenario 2}: A and B have created the chemical composition of an organic, temperature-regulating soil. A and B need additional funding to produce an
amount large enough to be tested on a greater scale \& form a viable business. A and B’s friend C is willing to invest 2M for a 30\% ownership interest, which A and B agree to. If all goes well, the three want to sell the soil’s chemical composition to a biotech company.}
    \label{fig:sankey2}
\end{subfigure}
\caption{\small \textbf{Comparing different human workflows}: In Scenario 1 where the client was uncertain about next steps, the workflow became more exploratory and recursive, with the student circling back to earlier tasks to clarify gaps. In contrast, the workflow was more streamlined in the more straightforward scenario 2, focusing on next steps and formalizing recommendations.
} 
\label{fig:sankey}
\end{figure}
Some scenarios mentioned in the main paper, along with their descriptions are provided below. Figure \ref{fig:sankey} highlights the difference in human workflows for different scenarios. 
\begin{quotation}
\noindent\scriptsize \textbf{scenario 10}: 

\noindent\scriptsize Title: Three-Person Coffee Truck with Entity Formation Contingency
Context:
Three friends, A, B, and C, want to form a coffee shop out of a food trailer that A purchased while the three were college roommates. They had informally used the trailer to sell coffee products on campus, but would like to see if they could form a real business. The key to their popularity has been A’s importation of Hawaiian coffee beans, which help the trio brew delicious coffee drinks but has proven to be an expensive business decision. If they form an entity, they might want to see whether they could have investors supply some extra money to help them cover the cost of importing the beans. They would plan on giving the investors some kind of profit interest in the business in exchange for their help, but aren’t too familiar with how that would work.

In addition, this venture has largely been A’s project. B and C aren’t quite as invested, as they have more time-consuming careers of their own. The group’s other friends, D and E, however, have kept in touch with A and have expressed interest in joining the business in place of B and C. B and C might be open to this as well, but will need some time to think. A must also later decide if he wants anyone involved with the business to be able to come and go as they please, or if he wants a firmer commitment from anyone involved with the truck (either B and C or D and E) to devote more of their time and resources to it.
Details:
1. Three-person
2. Sale of Goods
3. Personal Property
4. Entity Selection
Complexity Tags:
1. Memo needed
2. Diverging Interests of Clients
8. Multiple Sessions Needed for Entity Choice
Complexity: Medium \\

\noindent \textbf{scenario 13}: 

\noindent\scriptsize Title: One Person Ice Cream Maker
Context:
E has recently developed a process for making premium homemade ice cream with locally-sourced ingredients. The result has been a hit: E’s flavors are wide-ranging, unique, and high-quality. E has started by sharing the products with neighbors, who have suggested that he sell it for profit. E has not yet done so, but might be interested. He is wondering how to best proceed by beginning with local sales and going from there.
Details:
1. One person
2. Inexperienced client
Issues:
1. Formation
2. Branding
3. IP \\

\noindent \textbf{scenario 15}: 

\noindent\scriptsize Title: Lake Bed and Breakfast
Context:
M and A want to operate a bed and breakfast on the lake where they keep a cabin – just outside a small town in the northern, rural part of their home state. They intend to run a relatively small operation: the building they have picked out is a four bedroom house that can accommodate up to ten people comfortably. It is in relatively good shape, but would benefit from a handful of basic renovations. M and A will cook breakfasts, provide housekeeping services, and lead hikes for guests in the surrounding area. They will also provide kayaks, stand up paddleboards, and jet skis for rent, all of which are personally owned by the couple. They have a sufficient (i.e. covering everything) liability waiver in place which guests agree to upon booking their stay. Per an informal agreement, M and A have also received some funding from the nearby town’s chamber of commerce in exchange for recommending other local businesses to their guests. They’re interested in forming a business to separate their personal assets from that of the business.
In addition, they’d like to pass the business to their three children (ownership divided equally among them), and want to add them to the business now, each with partial ownership (M and A each holding a 25\% ownership). They would also like a provision in any operating document to specify that, should anyone seek to sell their ownership interest, the remaining members must get the first offer or otherwise unanimously consent in writing to the transfer.
Details:
1. Two Person
2. Services
3. Varying Liability
4. Transfer of Business
Issues:
1. Entity Formation
2. Addition of Members
3. Restrictions on Interest Transfer
Complexity Tags:
1. Memo needed
2. Diverging interests of clients \\

\noindent \textbf{scenario 17}: 

\noindent\scriptsize Title: Fishing Education
Context:
A and B are each avid anglers and seek to grow the sport within their area, which is largely urban. They seek to teach fishing skills—ranging from basic to advanced—to both children and adults, through events held at parks, lakes, and rivers nearby. They envision weekly or monthly workshops where they educate participants on a particular skill—casting, knot tying, fish landing with a net, and more. Signup for the workshops is free. A and B encourage participants to bring their own fishing gear, but plan on contracting with a local sporting goods store to purchase bulk orders of tackle at a discounted rate. Because of that, they have decided to explore forming a business entity.
Neither A nor B are particularly familiar with small business structure or ownership, and don’t seek to profit from the business. They would like, however, to minimize any formalities associated with owning a business. In addition, they would like to raffle off fishing apparel at some of the workshops, and are ok using their own money to buy these items for the first few instances. If their workshops are popular and they consistently have high attendance, they would like to borrow money or obtain outside funding to keep this practice up.
Overall, they want to learn more about the function of any documents used to start the business, and potentially review anything that’s written for them before they officially get to work.
Details:
1. Three Person
2. Low Liability
3. Informal Structure
Complexity Tags:
3. Follow-up for Operating Agreement Review
7. Inexperienced Client \\

\noindent \textbf{scenario 18}: 

\noindent\scriptsize Title: Shared Workspace and Kitchen
Context:
P is looking to bring together other businesses in his community by offering a space that functions as both a commercial kitchen and shared workspace, which can be leased by individuals or other entities. P has already formed an LLC for this purpose. The leases may be by the hour or by the day, and can be for the kitchen, spots in the workspace, or for the entire property. To minimize out-of-pocket expenses, P decides to obtain financing from a local bank for the necessary remodeling. In addition to the normal rentals, P has two other potential uses associated with the building. One is a regularly-occurring food truck fair that will feature local, newly-formed food trucks. It will take place on the building’s property, in its parking lot. The second is a temporary restaurant featuring one of P’s former business partners (B) who has become a chef. The restaurant will have the exclusive use of the space for ten weeks of the subsequent summer, and will be a collaborative effort between the two entities. P will be involved as well, as he and B had originally met while working together in another restaurant.
Details:
1. One Person
2. Lease agreement needed
Complexity Tags:
6. Well-prepared client \\

\noindent \textbf{scenario 19}: 

\noindent\scriptsize Title: Refurbishing Hockey Equipment
Context:
L accepts used hockey equipment—generally, skates and sticks—and repairs it, sometimes to give back to those who donate it, and other times for resale, if the donating party has no further use for it. The work is mostly done in his home, and varies based on how much equipment he has at a given time. He’s considering implementing the option of selling the resale equipment to a local hockey equipment store, mostly depending on if he keeps getting a supply of broken sticks—the piece of equipment most common on the resale market. Though buyers at these stores are aware of the potential defects of a refurbished hockey stick, the stores selling them generally provide no warranty on them, effectively relieving them of liability. L is confident—and correct—that there are no huge liability risks associated with selling refurbished sticks, but might want some protection—as well as a more official entity—to do business with the stores, and the consumers he sells to on the side. \\

\noindent \textbf{scenario 20}: 

\noindent\scriptsize A, B, and C have all worked together at a health clinic in the Twin Cities for the past 11 years. A and B are both physicians who have client relationships that would follow them from the current clinic to the one they are creating. Even though they will not be able to directly contact the clients from their current practice due to a non-compete agreement, they are well known and respected in the community and expect that many clients will follow them without proactive contact. C is an administrative professional who has practically run the current practice for the last 20 years and will be vital to the day-to-day operations of the new entity. C will be contributing all of the labor to running the new entity, including hiring support staff, scheduling appointments for the doctors, etc. The three have identified a suitable location for the new practice and may either purchase or lease it long-term depending on how the negotiations go. During the conversation, it should become clear that C expects an equal share of the new entity and is also worried about having decision-making power after formation. C does not have as much money to contribute as the doctors do and will be contributing mostly labor. A and B are somewhat dismissive of C’s value and expect to be “running the show” themselves because of their extensive training as doctors, etc. They find C indispensable, but not necessarily worthy of equal ownership in the new entity.
Tags associated : 
1.Three-Person (just two clients needed for simulation)
2.Financial Considerations
3. Entity Growth
Complexity Tags
1. Memo Needed
Nuance Tags 
4. Follow-up for Entity Choice
\end{quotation}

\begin{table}[ht]
\centering
\scriptsize
\caption{\small \textbf{Diversity and Complexity of Scenarios} split by Category}
\resizebox{\linewidth}{!}{%
\begin{tabular}{|>{\raggedright\arraybackslash}p{3.5cm}|
                >{\centering\arraybackslash}p{1.1cm}|
                >{\centering\arraybackslash}p{3cm}|
                >{\raggedright\arraybackslash}p{5.1cm}|}
\hline
\textbf{Category} & \textbf{Count} & \textbf{Domains} & \textbf{Complexity Summary} \\
\hline
Nonprofit / Mission-Driven & 3 & Health; Education & Simple setup; compliance burden; no profit motive \\
\hline
Startups / Entrepreneurial & 3 & Food; Agriculture & Growth-focused; needs funding, IP setup, and lease structuring \\
\hline
Family / Lifestyle Business & 3 & Hospitality; Food; Sports Equipment & Often solo-run; asset or succession planning adds complexity \\
\hline
Professional Practices & 1 & Professional Services & High regulatory demands; ownership and ethical constraints \\
\hline
\end{tabular}
}
\label{tab:scenario-diversity}
\end{table}

\begin{table}[ht]
\centering
\scriptsize
\caption{\small Calculated \textbf{Metrics for Business Entity} Formed Per Scenario} 
\resizebox{\linewidth}{!}{%
\begin{tabular}{|>{\raggedright\arraybackslash}p{1.8cm}|
                >{\centering\arraybackslash}p{1.4cm}|
                >{\centering\arraybackslash}p{1.4cm}|
                >{\centering\arraybackslash}p{2.4cm}|
                >{\centering\arraybackslash}p{2.2cm}|
                >{\centering\arraybackslash}p{2.6cm}|}
\hline
\textbf{Metrics} & \textbf{C-corp} (1) & \textbf{Nonprofit} (1) & \textbf{LLC} (6) & \textbf{No OA} (2) & \textbf{Total Avg} (10) \\
\hline
Agreement length & 5815.00 & 5071.00 & 3704.67 ± 1158.00 & -- & 4139.25 ± 1282.54 \\
\hline
Memo length & 262.00 & 514.00 & 380.17 ± 183.06 & 577.50 ± 207.18 & 421.20 ± 183.61 \\
\hline
Tools used & 43.00 & 35.00 & 77.67 ± 49.29 & 30.50 ± 20.51 & 60.50 ± 43.58 \\
\hline
\end{tabular}
}
\label{tab:business-entity-metric}
\end{table}

\subsubsection{LLM Workflow Generation}
\label{sec:llm_prompt}

\paragraph{LLM Plan Graph Prompt}
\label{sec:llm_plan_prompt}

The prompt we used for an AI generated task plan for the entity formation scenario is given in Figure \ref{fig:llm_plan_prompt}. This is analogous to the human task plan. 

\paragraph{LLM workflow generation prompt}
\label{sec:llm_workflow_prompt}
\begin{figure}[t!]
\centering
\begin{minipage}[t]{0.95\columnwidth}
    \begin{lstlisting}
You are an experienced lawyer specializing in entity formation at a reputable American law firm. Your task is to create a structured, high-level reasoning plan outlining the steps you would consistently follow for any business formation scenario, from initial client consultation through the final drafting of the Operating Agreement.

To inform your thinking, consider these illustrative business scenarios:

<business_scenarios>
@{BUSINESS_SCENARIOS}@
</business_scenarios>

These scenarios are provided as examples to guide your reasoning. ....

1. Identify key components of entity formation
.
.

It's okay for this section to be quite long to ensure a thorough analysis.

After completing your analysis, ..

1. Clearly number the main tasks (e.g., "1. Task Name").
.
.
.

Example format [Only for demonstration purposes]:

1. Main Task One
   1.1. Sub-task A
   1.2. Sub-task B [Parallel/Concurrent]
      1.2.1. Sub-sub-task i
      1.2.2. Sub-sub-task ii
   1.3. Sub-task C
.
.

Important: Create your plan based on your professional....
    \end{lstlisting}
        \caption{LLM Plan Generation Prompt}
        \label{fig:llm_plan_prompt}
    \end{minipage}
\end{figure}

The prompt we use to generate the AI workflow is given in Figure \ref{fig:llm_workflow_prompt}. This is analogous to the human workflow. A snippet of the output from this prompt is shown in Figure \ref{fig:llm_cot_out1}. The output of the LLM along with the next step suggested by it is fed as context to the LLM again to generate the details of the next step. To start the execution of a particular scenario by the LLM, we provide a prompt similar to Figure \ref{fig:llm_workflow_prompt} but without the additional context as there are no prior steps to be input. 

\begin{figure}[t!]
    \centering
    \begin{minipage}[t]{0.95\columnwidth}
        \begin{lstlisting}
        
You are an AI assistant tasked with generating realistic interactions related to small business formation. Your goal is to create either a dialogue between a lawyer and clients or a description of a lawyer's actions, based on a specific step in a high-level plan for business formation.

First, review the following information:

1. High-level plan for business formation:
<high_level_plan>
@{task}@
</high_level_plan>

2. Specific scenario for this business formation:
<scenario>
@{scenario}@
</scenario>

3. Previous steps and context [Note: The steps at the top are the earliest ones, and the ones at the bottom are the latest ones.]:
<previous_steps>
@{context}@
</previous_steps>

4. Suggested step for this turn:
<suggested_step>
@{suggested_step}@
</suggested_step>

Your task is to generate either a conversation or a description of the lawyer's actions based on the suggested step. Follow these instructions carefully:


1. Analyze the current step and plan the interaction:
<step_analysis>
.
.
</step_analysis>

2. Assess the lawyer's knowledge at this stage:
<knowledge_assessment>
.
.
</knowledge_assessment>

3. Generate the interaction:
If the step involves both lawyer and clients:
<conversation>
Lawyer: [Lawyer's dialogue]
Client: [Client's dialogue]
[Continue the conversation as needed]
</conversation>

If the step involves only the lawyer:
<lawyer_action>
[Describe the lawyer's actions, thought process, and any documents or research they might be working on]
</lawyer_action>

4. Suggest the next step:
<next_suggested_step>
[Specify the next step from the plan]
</next_suggested_step>


Remember:
- Ensure that your dialogue or action description is relevant and tailored to the given business context.
.
.

### Answer  
    \end{lstlisting}
        \caption{LLM Workflow generation prompt}
        \label{fig:llm_workflow_prompt}
    \end{minipage}
\end{figure}

\begin{figure}[t!]
    \centering
    \begin{minipage}[t]{0.95\columnwidth}
        \begin{lstlisting}
$Step 1$
<step_analysis>
.
.
</step_analysis>

<knowledge_assessment>
   .
   .
</knowledge_assessment>

@<lawyer_action>@
The lawyer blocks off time to speak with two experienced colleagues in the firm's business-formation group. In a confidential setting, the lawyer outlines the broad outlines of the matter - three individuals pooling resources for a healthcare practice, with one non-physician demanding equal equity but bringing primarily sweat equity. The lawyer highlights concerns about diverging interests and invites feedback on:

- Whether representing all three individuals raises insurmountable conflicts of interest - or if representing the entity might be more suitable.  
- How other attorneys have handled situations where a key administrative partner wants a significant stake despite limited capital.  
- Applicable state regulations or ethical considerations related to professional service entities mixing physician and non-physician ownership.  

One colleague suggests reviewing the state's professional corporation and LLC statutes to see if non-physician ownership holds any special limitations. Another colleague emphasizes the sensitivity of the doctors' non-compete constraints and advises clarifying how patient follow-up will occur. Both recommend the lawyer consider sending each client for independent counsel or at least addressing the conflict-of-interest disclosures in writing.

The lawyer takes detailed notes, planning to incorporate the colleagues' suggestions into the next steps, including verifying local healthcare licensing rules on non-doctor ownership, reevaluating how to proceed with joint representation (if feasible at all), and preparing to speak with the clients about potential alternative structures. The lawyer concludes by thanking the colleagues and scheduling a follow-up to share any further updates once new information is gathered.
@</lawyer_action>@

@<next_suggested_step>@
2.1 Think about recommendations
@</next_suggested_step>@

$Step 2$
...
@<lawyer_action>@
The lawyer sits down at their desk and begins organizing the considerations drawn from both legal research and previous colleague consultations. They jot down a list of possible recommendations that address ownership structure, management authority, and compliance with state professional practice requirements.

First, the lawyer reviews the state laws that govern professional service entities. They note that some states allow a non-physician to hold a partial stake in a medical practice, but with restrictions on certain decision-making powers related to clinical care. The lawyer sketches out possible corporate forms (e.g., forming a professional limited liability company versus a professional corporation) and highlights how each structure might impact management rights and liability coverage. They also reflect on conflict-of-interest rules and the duty to inform the potential co-owners about the lawyer's role.

Next, the lawyer drafts a series of bullet points for each recommended approach to ownership and governance:

.
.

Finally, the lawyer places a star next to the key question of how to handle the potential conflict of interest in representing all three. The lawyer contemplates whether to send short disclaimers or have an upfront conversation with each client. The next step, the lawyer decides, is to weigh the pros and cons of each ownership approach and prepare a concise outline to discuss with the clients.
@</lawyer_action>@

@<next_suggested_step>@
2.1.1 Consider counter-arguments/pros and cons
@</next_suggested_step>@
    \end{lstlisting}
        \caption{LLM Execution Outputs showing 2 consecutive steps generated}
        \label{fig:llm_cot_out1}
    \end{minipage}
\end{figure}

\subsection{Additional Analyses and Results}
\subsubsection{Human vs LLM Plans}
\begin{figure}[htbp]
    \centering
    \begin{subfigure}{\textwidth}
        \centering
        \includegraphics[width=\textwidth,keepaspectratio]{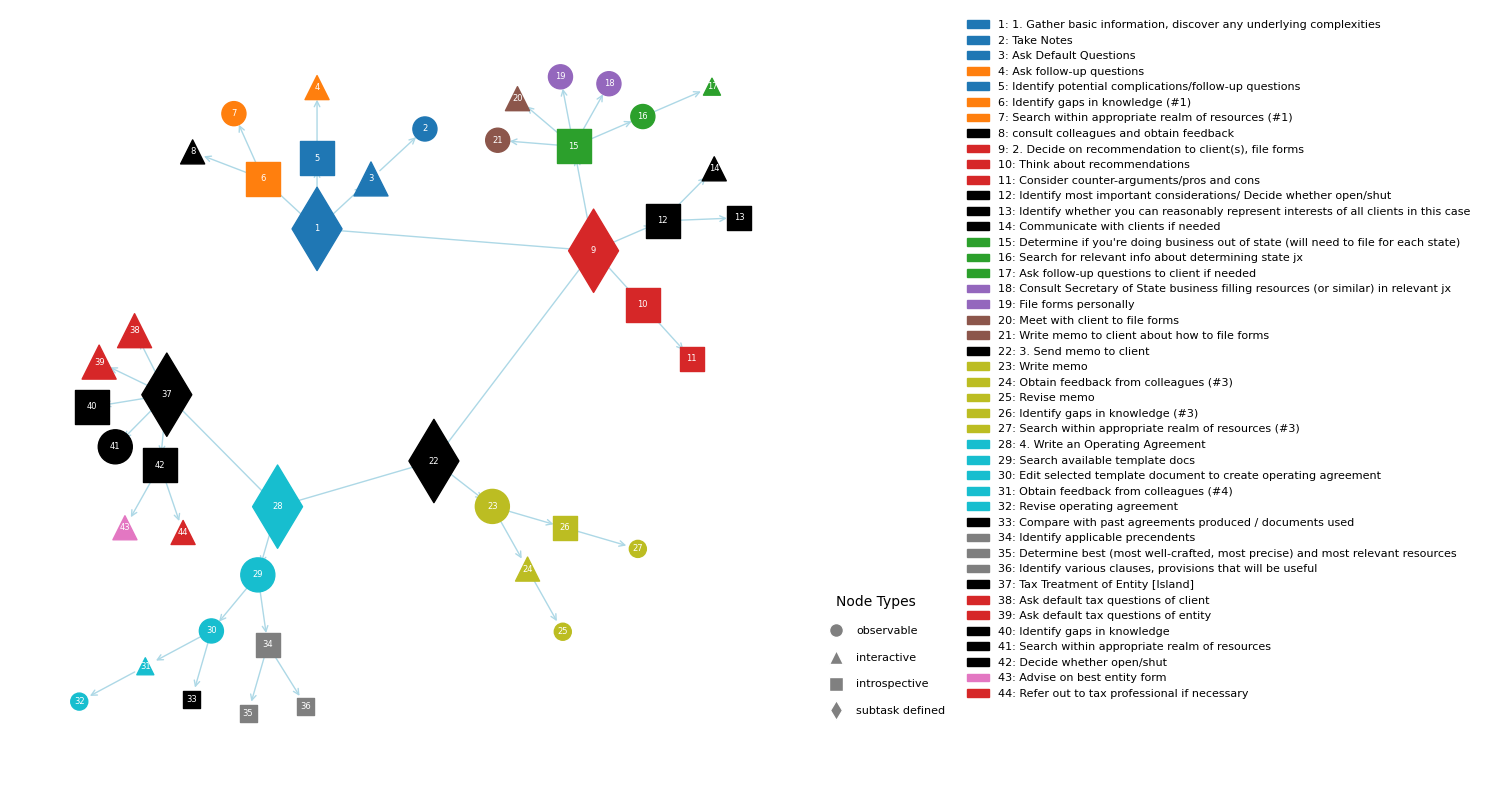}
        \caption{Human Task Plan,  color mapped to LLM Task Plan}
        \label{fig:human_plan}
    \end{subfigure}

    \vspace{0.5cm} 

    \begin{subfigure}{\textwidth}
        \centering
        \includegraphics[width=\textwidth,keepaspectratio]{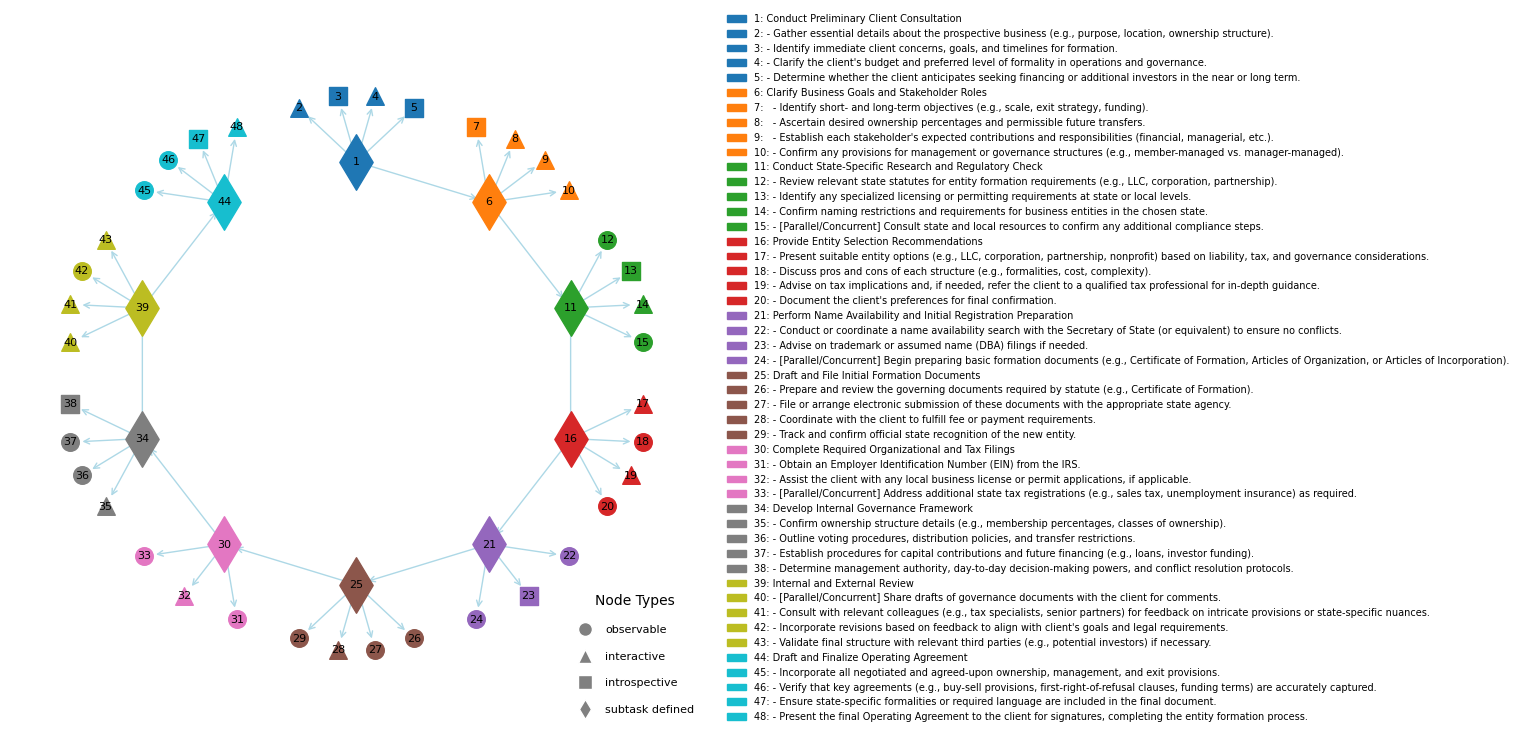}
        \caption{LLM Task Plan}
        \label{fig:llm_plan}
    \end{subfigure}

    \caption{This figure shows the mapping of all subtasks from the LLM Task Plan onto the Human Task Plan. Black nodes in the Human Task Plan represent steps that exist in the Human Task Plan but have no corresponding match in the LLM Task Plan, indicating that LLM-generated plans omit these steps. Additionally, multiple LLM subtask nodes often map under a single human subtask node, suggesting that human plans are more modular and hierarchically structured than LLM plans.}
    \label{fig:human_llm_plan}
\end{figure}

Figure \ref{fig:human_llm_plan} shows that multiple LLM subtasks often map to a single subtask in the human plan, indicating coarser granularity. Black nodes in the human plan represent tasks not captured in the LLM plan, for example ``consulting colleagues and obtaining feedback'' or ``identifying gaps in knowledge'' are common human steps which are not accounted for by the LLM. These patterns reinforce a key distinction: \textit{human plans emphasize modularity and depth, while LLM plans prioritize surface-level breadth}.

\subsubsection{Workflow Monitor}

\begin{table}[tbp]
  \centering
  \vspace{-4mm}
  {\small
  \renewcommand{\arraystretch}{1.6}
  \setlength{\tabcolsep}{8pt}
  \begin{tabular}{|p{3.5cm}|c|}
    \hline
    \textbf{Flag (Deviation Range)} & \textbf{Percentage (\%)} \\
    \hline
    On track (0–0.5) & 67.82 \\
    \hline
    Needs review (0.5–2) & 14.78 \\
    \hline
    Off track ($>$2) & 17.39 \\
    \hline
  \end{tabular}
  \caption{\small Z-score distributions show that nearly 70\% of LLM-generated steps align with human-like sequence planning.}
  \label{tab:deviance_band}
  }
\end{table}



The workflow monitor is a data-driven model trained to detect deviations from human-like reasoning in either other human or LLM-generated workflows. We trained a Llama-3.2-1B-Instruct model \citep{grattafiori2024llama3herdmodels} on eight human-generated workflow scenarios to act as a step-level classifier. Given prior context, the model predicts the most probable next step and uses perplexity to estimate how typical a step is, with lower perplexity indicating closer alignment with human behavior. We hold out 2 human workflows as a test set along with 10 LLM generated workflows. 

Next, we define a \textbf{deviance score} as the z-score of a conditional step-wise perplexity (perplexity of only the new generated next-step) relative to the human distribution:
\[
\text{Deviance Score} = \frac{\text{step perplexity} - \mu_{\text{human perplexity}}}{\sigma_{\text{human perplexity}}}
\]

where \(\mu_{\text{human perplexity}}\) and \(\sigma_{\text{human perplexity}}\) represent the mean and standard deviation of step-wise human perplexities across the human sequences upon which the model is trained. This represents the deviations relative to the distribution of human behavior. Steps with deviance scores \textit{beyond two standard deviations are flagged as unlikely}, while those near zero suggest strong alignment. Moderate deviations may signal ambiguity or areas that require human input. Rather than enforcing correctness, this monitor provides soft signals to prompt reflection and revision, helping AI systems stay on track while supporting flexible, context-sensitive legal reasoning. Results (Table ~\ref{tab:deviance_band}) show that \textit{67.8\% of LLM-generated steps align closely with human behavior}, while others fall into moderate and high deviation bands.

\subsubsection{Examples of Workflow Outputs}
For each scenario, in addition to a human drafting the appropriate agreement, the LLM is also used to generate an agreement for the scenario in order to enable comparison. The prompt used to assist the LLM in generating the agreement can be found in Figure \ref{fig:llm_oa}.
\begin{figure}[t!]
\centering
\begin{minipage}[t]{0.95\columnwidth}
    \begin{lstlisting}
You are a lawyer. Your role is to analyze the provided business scenario, determine the most suitable business entity to form for your clients, and select the most appropriate template from a given list to draft a complete, legally enforceable governing document.

### Instructions
Analyze the given business scenario, determine the most suitable business entity type, choose an appropriate template from the provided list, and draft a complete, legally binding document.
### Scenario Title
{TITLE}
### Scenario Description
{DESCRIPTION}
### Available Templates
Template 1
[...]
Template 2
[...]
### Business Entity Recommendation, Template Selection and Writing an Agreement


    \end{lstlisting}

        \caption{LLM Agreement Generation Prompt}
        \label{fig:llm_oa}
    \end{minipage}
\end{figure}

LLM-generated and human-generated agreements for these two cases are described:

(1) A case where the \textit{LLM agreed} with the human on the same business entity recommendation and \textit{selected the same template}. Scenario considered :  Simulation 19 - Refurbishing Hockey Equipment (Appendix A.2). Snippets of this human-generated and LLM-generated agreement are shown in Figure \ref{fig:simulation_19_human_oa} and Figure \ref{fig:simulation_19_llm_oa}.
\begin{figure}[t!]
\centering
\begin{minipage}[t]{0.95\columnwidth}
    \begin{lstlisting}
OPERATING AGREEMENT OF MADSTICKS LLC.
 
This Operating Agreement (this ``Agreement'') is entered into by and between Brendan Madson [...]

Name.  The name of the limited liability company is Madsticks LLC [...]


Assumed Name.  The Company may do business under an assumed name from time to time, as provided under Chapter 322C of the Minnesota Statutes [...]

Purpose.  The purpose and character of the business of the Company shall be to undertake and carry on any lawful business, purpose [...]

Operating Agreement.  This Agreement constitutes the ``Operating Agreement'' of the Company as defined in Section 322C.0102, [...]

Principal Place of Business.  The principal business office of the Company will be located at 123 Hockey Lane, St. Paul, [...]

Registered Office and Registered Agent.  The location of the registered office and the name of the registered agent (if any) of the Company [...]

No Personal Liability.  The liability of the Sole Member shall be limited to the fullest extent permitted by law [...]

Capital Contributions; Membership Interest in the Company.  The Sole Member's capital contribution is set forth on the attached Schedule A [...]

Additional Contributions.  The Sole Member is not required to make any additional capital contributions to the Company [...]

    \end{lstlisting}
        \caption{\small Operating Agreement drafted by Human (\textit{LLM agreed with the human} on the same business entity recommendation and selected the same template.)}
        \label{fig:simulation_19_human_oa}
    \end{minipage}
\end{figure}

\begin{figure}[t!]
\centering
\begin{minipage}[t]{0.95\columnwidth}
    \begin{lstlisting}
# OPERATING AGREEMENT
OF
L's Refurbished Hockey Equipment LLC

This Operating Agreement (this ``Agreement'') is entered into by and between L [Last Name] (the ``Sole Member'') and L's Refurbished Hockey Equipment LLC, a Minnesota limited liability company (the ``Company''), as of the Effective Date (as defined below)[...]

1. **Name.** The name of the limited liability company is L's Refurbished Hockey Equipment LLC.

2. **Assumed Name.** The Company may do business under an assumed name from time to time, as provided under Chapter 322C of the Minnesota Statutes[...]

3. **Purpose.** The purpose and character of the business of the Company shall be to undertake and carry on the business of refurbishing used hockey equipment[...]

4. **Operating Agreement.** This Agreement constitutes the ``Operating Agreement'' of the Company as defined in Section 322C.0102,[...]

5. **Principal Place of Business.** The principal business office of the Company will be located at [Your Home Address], or such location as may hereafter be determined by the Sole Member.

6. **Registered Office and Registered Agent.** The location of the registered office and the name of the registered agent (if any) of the Company[...]

7. **No Personal Liability.** The liability of the Sole Member shall be limited to the fullest extent permitted by law[...]

8. **Capital Contributions; Membership Interest in the Company.** The Sole Member's capital contribution is set forth on the attached Schedule A[...]

9. **Additional Contributions.** The Sole Member is not required to make any additional capital contributions to the Company.[...]


    \end{lstlisting}
        \caption{\small Operating Agreement drafted by LLM (\textit{LLM agreed with the human} on the same business entity recommendation and selected the same template.)}
        \label{fig:simulation_19_llm_oa}
    \end{minipage}
\end{figure}

(2) A case where the \textit{LLM agreed} with the human on the same business entity recommendation but \textit{selected a different template} than the human. Scenario considered : Simulation 15 - Lake Bed and Breakfast. The description of the simulation can be found in Appendix A.2. Snippets of this human-generated and LLM-generated agreement are shown in Figure \ref{fig:simulation_15_human_oa} and Figure \ref{fig:simulation_15_llm_oa}.
\begin{figure}[t!]
\centering
\begin{minipage}[t]{0.95\columnwidth}
    \begin{lstlisting}
OPERATING AGREEMENT 
[COMPANY] 

This  Operating  Agreement  (this  ``Agreement'')  of  Aunt  Doris's  B&B,  L.L.C.,  a Minnesota limited liability company (the ``Company''), is entered into as of December 30, 2024  (the ``Effective Date'') by and among the Company, Robin, and Daniel (each a ``Member'' and collectively the ``Founding Members''), and any other person or entity that, after the date above, becomes a Member of the Company in accordance with the terms of this Agreement.  

ARTICLE 1 
Organizational Matters 

Section 1. 

Name. The name of the Company is Aunt Doris's B&B, L.L.C. 

Section 2.   Assumed Name; Business Name. The Company may do business under an 
assumed  name  from  time  to  time,  as  provided  by  the  Minnesota  Revised  Uniform  Limited 
Liability  Company  Act,  Minn[...]

Section 3. 

Principal  Office.  The  principal  office  of  the  Company  is  located  at 
[ADDRESS],[...]

Section 4. 

Registered  Office;  Registered  Agent.  The  registered  office  and  agent for 
service  of  process  on  the  Company,  in  the  State  of  Minnesota,[...]

Section 5.  

Purpose; Powers. 

(a)  The  purpose  of  the  Company  is to engage in any lawful business purpose or 
activity in accordance with the MN RULLCA[...]

    \end{lstlisting}
        \caption{\small Operating Agreement drafted by Human (LLM agreed with the human on the same business entity recommendation but \textit{selected a different template} than the human)}
        \label{fig:simulation_15_human_oa}
    \end{minipage}
\end{figure}

\begin{figure}[t!]
\centering
\begin{minipage}[t]{0.95\columnwidth}
    \begin{lstlisting}
OPERATING AGREEMENT
OF
Lake Bed and Breakfast LLC

TABLE OF CONTENTS

Article 1 Definitions 	1  
Article 2 Formation 	4  
Article 3 Capital Contributions 	5  
Article 4 Allocations of Profits and Losses; Distributions 	6  
Article 5 Management 	7  
Article 6 Books and Records; Tax Matters 	11  
[...]

THE LIMITED LIABILITY COMPANY INTERESTS (OR ``UNITS'') OF THE COMPANY DESCRIBED IN AND GOVERNED BY THIS AGREEMENT HAVE NOT BEEN REGISTERED UNDER THE SECURITIES ACT OF 1933, AS AMENDED, OR UNDER ANY APPLICABLE STATE SECURITIES LAWS. THE UNITS ARE RESTRICTED SECURITIES WITHIN THE MEANING OF RULE 144 PROMULGATED UNDER THE SECURITIES ACT OF 1933, AS AMENDED[...]

OPERATING AGREEMENT
OF
Lake Bed and Breakfast LLC

This OPERATING AGREEMENT (``Agreement'') is made this ______ day of _______, 202__, by and between M and A and their three children, collectively, the ``Members'' and each, individually, a ``Member''.

Recitals
The undersigned constitute all of the current Members of the Company.  
Each of the undersigned desires to enter into this Agreement, [...]
Agreement
In consideration of the foregoing and the mutual promises and agreements set forth below, the Members agree as follows:

### Article 1. Definitions

The terms defined in this Article 1 [...]
``Act'' means the Minnesota Revised Uniform Limited Liability Company Act [...]

``Additional Member'' means a Person who is admitted as a Member and issued a new Company Interest.

``Affiliate'' means, with respect to any Person, (i) any Person that directly or indirectly through one or more intermediaries controls or is controlled by or is under common control with the specified Person, [...]

    \end{lstlisting}
        \caption{\small Operating Agreement drafted by LLM (LLM agreed with the human on the same business entity recommendation but selected a \textit{different template} than the human)}
        \label{fig:simulation_15_llm_oa}
    \end{minipage}
\end{figure}

\begin{figure}[t!]
    \centering
    \begin{minipage}[t]{0.95\columnwidth}
        \begin{lstlisting}
@Formal Criteria@
    - Are all parts consistent with each other?
    - Are all the terms clearly defined?
    - Is the language of the agreement clear?
    - Are all revisions consistent throughout the document?
    - Does it meet the minimal standards of the operating agreement checklist? (See attached)

@Substantive Criteria@
    - Is this OA workable in practice?
    - Will the OA lead to disputes?
    - Does the OA capture the actual intent of the business?
    - Will this OA be difficult to exit or unwind?
    \end{lstlisting}
        \caption{Rubric for Evaluating Operating Agreements}
        \label{fig:eval_rub}
    \end{minipage}
\end{figure}

\begin{figure}[t!]
    \centering
    \begin{minipage}[t]{0.95\columnwidth}
        \begin{lstlisting}
## Answer these questions based on the following scenario description and its operating agreement. Provide a score out of 5 for each question, along with reasoning for the score.  

### Formal  

- Are all parts of the operating agreement internally consistent? (Score: X/5, Reasoning: ...) 

- Are all key terms clearly defined in the document? (Score: X/5, Reasoning: ...) 

- Is the language clear and precise? (Score: X/5, Reasoning: ...) 

### Substantive  
- Does the operating agreement reflect all customer requirements? (Score: X/5, Reasoning: ...) 

- Is the operating agreement practical and enforceable? (Score: X/5, Reasoning: ...) 

- Are there clauses that might create unnecessary disputes between stakeholders? (Score: X/5, Reasoning: ...) 

- Does the operating agreement capture the intended structure and goals of the business? (Score: X/5, Reasoning: ...) 

- Will this operating agreement make it difficult for stakeholders to exit or dissolve the entity if needed? (Score: X/5, Reasoning: ...) 
    
### Scenario Description  
Title: @{TITLE}@
Description: @{DESCRIPTION}@

### Operating Agreement}  
@{OA}@

### Answer  
    \end{lstlisting}
        \caption{LLM-as-Judge general questions evaluation prompt}
        \label{fig:llm_eval_prompt_part1}
    \end{minipage}
\end{figure}

\begin{figure}[t!]
\centering
\begin{minipage}[t]{0.95\columnwidth}
    \begin{lstlisting}
## Answer these questions based on the following scenario description and its operating agreement. Provide a score out of 5 for each question, along with reasoning for the score.  

### Formal  
- Does it follow the minimal standards of an operating agreement checklist? (Score: X/5, Reasoning: ...) 

#### Operating Agreement Checklist  
@{CHECKLIST}@ 

### Scenario Description  
Title: @{TITLE}@
Description: @{DESCRIPTION}@ 

### Operating Agreement  
@{OA}@

### Answer 
    \end{lstlisting}
        \caption{LLM-as-Judge LLC Checklist question evaluation prompt}
        \label{fig:llm_eval_prompt_part2}
    \end{minipage}
\end{figure}

\subsubsection{Evaluation of operating agreements}
Figure \ref{fig:eval_rub} describes the expert-defined evaluation rubric for the entity formation scenario's agreements generated. This rubric is further used to develop LLM-as-Judge evaluation prompts for the Agreement as shown in figure \ref{fig:llm_eval_prompt_part1} and figure \ref{fig:llm_eval_prompt_part2}. These prompts are use the evaluation rubric.

\end{document}